\ifdefined\XeTeXversion\else\ifdefined\directlua\else\pdfoutput=1\fi\fi
\documentclass[11pt]{article}
\PassOptionsToPackage{hyphens}{url}
\usepackage[numbers]{clain}
\usepackage{fvextra}
\usepackage[bottom]{footmisc}
\definecolor{visualblue}{HTML}{193F35}
\definecolor{cultureorange}{HTML}{B76142}
\definecolor{softblue}{HTML}{E8EFE9}
\definecolor{softorange}{HTML}{F8F4EC}
\newcolumntype{Y}{>{\raggedright\arraybackslash}X}
\hypersetup{
  pdftitle={Polytropos: What Do Vision-Language Model Descriptions Make of a Cultural Object?},
  pdfauthor={Yuechen Jiang, Yuyan Wang, Mohsinul Kabir, Stavroula Golfomitsou, Konstantinos Arvanitis, Sophia Ananiadou}
}
\DeclareRobustCommand{\PolytroposBrand}{%
  \texorpdfstring{%
    \mbox{\fontencoding{T1}\fontfamily{qbk}\fontseries{b}\fontshape{it}\selectfont
      \textcolor{clainink}{Poly}\textcolor{clainred}{tropos}}%
  }{Polytropos}%
}
\title{\PolytroposBrand{}: What Do Vision-Language Model Descriptions Make of a Cultural Object?}
\date{22 November 2026}
\reportlabel{Technical Report}
\author{%
Yuechen Jiang\textsuperscript{1}, Yuyan Wang\textsuperscript{1},
Mohsinul Kabir\textsuperscript{1}, Stavroula Golfomitsou\textsuperscript{2},
Konstantinos Arvanitis\textsuperscript{1}, Sophia Ananiadou\textsuperscript{1,3}%
}
\affiliation[1]{University of Manchester}
\affiliation[2]{Getty Conservation Institute}
\affiliation[3]{ELLIS Manchester}
\correspondence{}
\metadata[Hugging Face]{\href{https://huggingface.co/datasets/Carolyn-Jiang/Culture-Ground-Description}{huggingface.co/datasets/Carolyn-Jiang/Culture-Ground-Description}}
\metadata[Website]{\href{https://carolyn-jiang.github.io/Polytropos/}{carolyn-jiang.github.io/Polytropos/}}

\begin{document}
\raggedbottom
\begin{abstract}
Vision-language models have become a new kind of describer, writing about almost anything people see, and their descriptions now shape how cultures are known by people and by other models. Descriptions of cultures tend to turn particular things into types, and what a description loses shows what its describer lacks. We introduce \PolytroposBrand{}, a benchmark that measures what model descriptions of cultural objects keep of the particular object by turning each description back into an image and comparing it with the original photograph. On 374 objects from the Metropolitan Museum of Art and the Getty, with 1,870 descriptions from five models and 230 written by curators, reconstructions are recognised as the recorded culture more often than the photographs, at 54.55\% to 59.09\% against 52.14\%, yet 16\% to 21\% of objects change outcome and only 0.49\% of reconstructions are identified as the specific object. On the same objects, models lose recorded cultures at a rate similar to curators but gain more. Model descriptions give each object many faces, each typical of its culture and almost none the object itself.
\end{abstract}
\maketitle

\section{Introduction}
\label{sec:introduction}
\begin{figure}[t]
\centering
\includegraphics[width=0.9\linewidth,height=0.39\textheight,keepaspectratio]{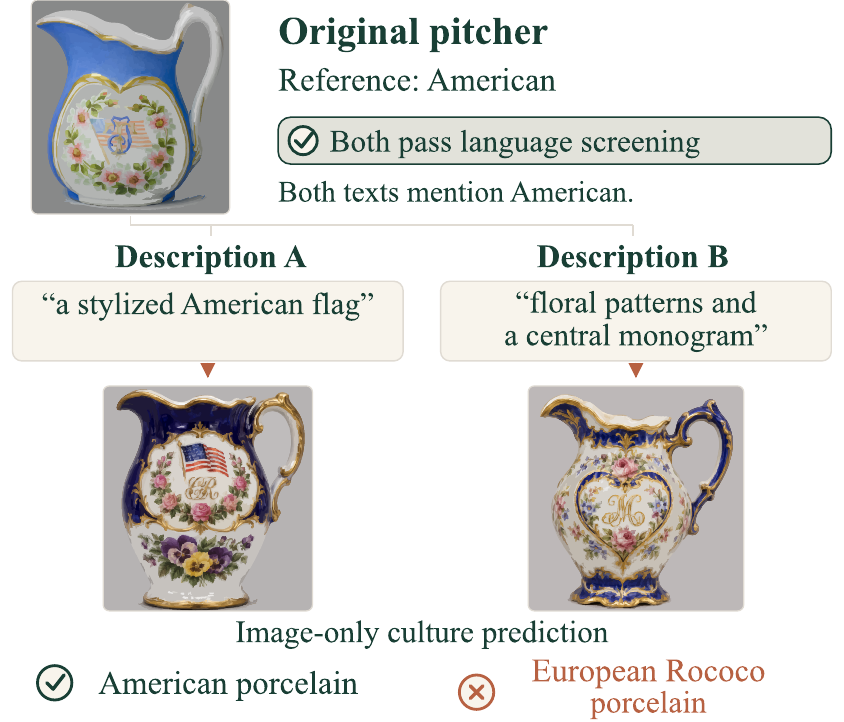}
\caption{Culture recovery differs despite language eligibility. A: Gemini-2.5-Flash-Lite; B: LLaVA-OneVision-1.5-8B. Images come from full unchanged descriptions; quotations are verbatim.}
\label{fig:task}
\end{figure}
\begin{table*}[t]
\centering\small
\setlength{\tabcolsep}{4pt}
\renewcommand{\arraystretch}{1.0}
\newcommand{\featureyes}{\textcolor{teal!75!black}{$\checkmark$}}
\newcommand{\featureno}{\textcolor{red!65!black}{$\times$}}
\caption{Comparison with related benchmarks, grouped as in Section~\ref{sec:related}. Prior benchmarks score model output against a reference, and human text serves as that reference. \PolytroposBrand{} judges each description against the photograph of the same object through a recogniser that never sees the text, and scores curators as a second describer.}
\label{tab:comparison}
\begin{tabularx}{\textwidth}{@{}>{\raggedright\arraybackslash}p{3.1cm}*{5}{>{\raggedright\arraybackslash}X}@{}}
\toprule
\textbf{Benchmark} & \textbf{Model output} & \textbf{Compared with} & \textbf{Judged by} & \textbf{Score reports} & \textbf{Human text as} \\
\midrule
\multicolumn{6}{@{}l}{\emph{Description and evaluation}} \\
COCO Captions~\cite{chen2015coco} & Caption & Crowd captions & Text overlap & Similarity & Reference, baseline \\
POPE~\cite{li2023pope} & Yes or no answer & Object annotations & Exact match & Accuracy & None \\
CycleReward~\cite{bahng2025cycle} & Caption & Source image & Image similarity & Reward & None \\
\midrule
\multicolumn{6}{@{}l}{\emph{Culture-aware models}} \\
CVQA~\cite{romero2024cvqa} & Multiple choice & Answer key & Exact match & Accuracy & Reference \\
CulturalVQA~\cite{nayak2024benchmarking} & Short answer & Human answers & LLM judge & Accuracy & Reference, baseline \\
\midrule
\multicolumn{6}{@{}l}{\emph{Museum and cultural heritage}} \\
iMet Collection~\cite{zhang2019imet} & Attribute labels & Museum labels & Label match & F2 score & None \\
Artpedia~\cite{stefanini2019artpedia} & Sentence match & Wikipedia sentences & Retrieval & Recall & Reference \\
GalleryGPT~\cite{bin2024gallerygpt} & Painting analysis & LLM analyses & Text overlap & Similarity & None \\
Attribution~\cite{jiang2026appear2meaningcrossculturalbenchmarkstructured} & Attribution fields & Museum record & LLM judge & Fields matched & None \\
\midrule
\textbf{\PolytroposBrand{} (ours)} & Description & Object's photo & Text-blind recogniser & Kept or replaced by type & Second describer \\
\bottomrule
\end{tabularx}
\end{table*}

Vision-language models have become a new kind of describer, writing about almost anything people see, including the objects of every cultural tradition~\cite{villaespesa2021comparison,bin2024gallerygpt,reshetnikov2025caption}. Their descriptions already enter catalogues, search results, and the training data of the next models~\cite{betker2023improving,perez2025telephone,shumailov2024ai}, and through them shape how present and future generations know cultures they will never see. Descriptions of cultures have long turned particular things into types~\cite{clifford1986writing,abulughod1991writing}. A describer does so where it knows a thing less well than its kind and fills the gap with the kind~\cite{huttenlocher1991categories,bartlett1932remembering}, which makes what a description loses a measure of what its describer does not know~\cite{bozeat2003duck}, a measure recognition cannot give, since knowing the kind suffices to name a culture. Do the descriptions of vision-language models keep the particular object, or only its kind?

\begin{figure*}[t]
\centering
\includegraphics[width=\textwidth]{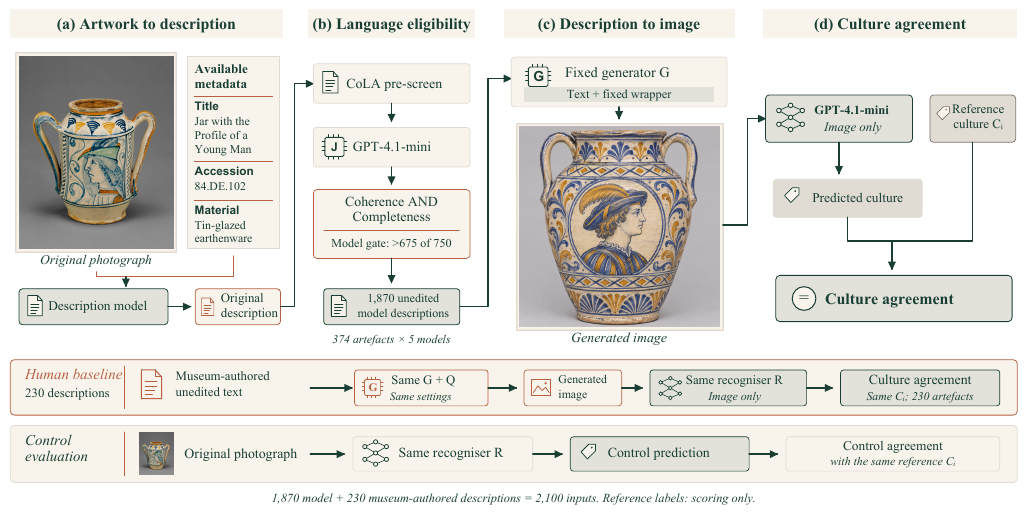}
\caption{Unedited model descriptions (1,870) and museum-authored baseline descriptions (230) share the same text-to-image generator and wrapper, image-only culture recogniser, and reference-based scoring; model--human comparisons use the same 230 artefacts. The separate original-photograph control assesses recognition error.}
\label{fig:workflow}
\end{figure*}

We propose \PolytroposBrand{}, the first benchmark that measures what model descriptions of cultural objects keep of the particular object (Table~\ref{tab:comparison}). A fluent description that names the right culture reads as accurate, even to a careful reader, so we turn each description back into an image and set it beside the original photograph, where what was kept, lost, or added becomes visible (Figure~\ref{fig:task}). The dataset contains 374 objects from the Metropolitan Museum of Art and the Getty, stratified by region and object type, with 1,870 descriptions from five vision-language models, selected from fifteen because their descriptions are well formed for more than 90\% of objects, and 230 museum-authored descriptions of the same objects as a human baseline. The task renders each unchanged description with a fixed text-to-image generator, and an image-only recogniser, which sees neither the description nor the source photograph, predicts the culture of the reconstruction (Figure~\ref{fig:workflow}). For evaluation, the same recogniser applied to the original photograph gives a paired reference, so each object keeps, loses, or gains its recorded culture under each description, model and museum descriptions are compared on the same objects, and an identification probe tests whether the particular object can still be recognised.

Across five models, reconstructions are recognised as the recorded culture more often than the photographs, at 54.55\% to 59.09\% against 52.14\%, yet 16\% to 21\% of objects change outcome, and only 0.49\% of reconstructions are identified as the specific object, against 2.67\% of photographs. On the 230 objects that also have museum descriptions, models lose a similar share of recorded cultures to curators, 9.13\% to 11.74\% against 11.74\%, but gain more, 10.87\% to 15.22\% against 10.87\%, moving objects toward what their culture typically looks like. Museum text also points to a particular object in 32.61\% of descriptions, against 7.83\% to 13.04\% for model text. The descriptions show how. They frame objects in broad traditions, such as a protective creature in East Asian tradition or the literati painting tradition, while turning a frontal pose into a back view and visible yellow into gold. The Americas are hardest. Of their 100 objects, 99 are recorded as American, yet only 25.0\% of photographs and 32.8\% of reconstructions are recognised as such, as American work is read through European styles. The new describer gives each object many faces, each more typical of its culture and almost none the object itself, as a describer that holds only the kind would~\cite{huttenlocher1991categories,jiang2026appear2meaningcrossculturalbenchmarkstructured}.

\section{Related Work}
\label{sec:related}
\paragraph{Computer vision: description and evaluation.}
Image description has developed from attention-based captioners~\cite{xu2015show,anderson2018bottom} to vision-language models that connect visual encoders with large language models~\cite{li2023blip2,liu2023llava}, and detailed-caption datasets support long, compositional descriptions~\cite{onoe2024docci,garg2024imageinwords}. Caption metrics measure agreement with reference captions~\cite{vedantam2015cider,anderson2016spice} or image-text compatibility~\cite{hessel2021clipscore}, hallucination benchmarks probe visual grounding~\cite{rohrbach2018object,li2023pope,guan2024hallusionbench}, and text-to-image evaluation checks prompt faithfulness through question answering or object detection~\cite{hu2023tifa,ghosh2023geneval}. Reconstruction-based evaluation ranks detailed captions by how faithfully a generator reconstructs the source image~\cite{bahng2025cycle}, and transmission-chain studies track how generated text changes as models reuse it~\cite{perez2025telephone}. These measures ask whether a description is accurate or similar to its source. We ask what it keeps of a particular object and what it replaces with a type.

\paragraph{Culture-aware NLP and multimodal models.}
Cross-cultural NLP distinguishes cultural from linguistic variation~\cite{hershcovich2022crosscultural}, and surveys organise how culture is represented and evaluated in language and multimodal models~\cite{adilazuarda2024culture,pawar2025cultural}. BLEnD and CulturalBench test everyday cultural knowledge across communities~\cite{myung2024blend,chiu2025culturalbench}, and CVQA, CulturalVQA and CultureVLM extend such tests to images~\cite{romero2024cvqa,nayak2024benchmarking,liu2025culturevlm}. Vision and generative models also cover cultures unevenly. Object recognition varies across regions~\cite{devries2019does,ramaswamy2023geode}, and text-to-image models reproduce cultural stereotypes and fail on culture-specific objects~\cite{kannen2024beyond,liu2024scoft}. These works test what models know about cultures or how they render them. We examine what a model's description of one object passes on, with the original photograph as a paired reference.

\paragraph{Museum and cultural-heritage AI.}
Museum datasets support attribute recognition in collections such as iMet~\cite{zhang2019imet}. SemArt pairs paintings with catalogue commentary, and Artpedia separates visual from contextual sentences~\cite{garcia2019semart,stefanini2019artpedia}. GalleryGPT and caption tuning for cultural heritage extend artwork description to multimodal language models~\cite{bin2024gallerygpt,reshetnikov2025caption}, and recent work tests whether such models can infer the catalogue attribution of a museum object from its image~\cite{jiang2026appear2meaningcrossculturalbenchmarkstructured}. These works evaluate descriptions or attributions against references. We turn descriptions back into images to see what they keep of the object, and compare them with descriptions written by curators.

\section{\PolytroposBrand{} Benchmark}
\label{sec:benchmark}
\PolytroposBrand{} turns each description back into an image to see what it keeps of the particular object. Its primary measure is agreement with the recorded museum attribution, and qualitative analysis examines object-specific detail. Figure~\ref{fig:workflow} summarises the information boundaries.

\subsection{Task Formulation}
\label{sec:task}
A description can keep two kinds of content. \emph{Instance-specific information} distinguishes one object from others of its kind, such as its patterns and layout, inscriptions, material, traces of making, form and condition. \emph{Type information} is what the object shares with recognisable conventions of its culture, which may also reflect stereotypes learned by models. Culture agreement measures whether the type survives, while identification probes and qualitative comparisons examine the instance. For artefact $i$ and description model $m$, let $I_i$ denote source photographs, $M_i$ catalogue metadata, and $y_{im}=f_m(I^{\mathrm{used}}_{im},M_i,P)$ the English description. The recorded images actually used satisfy $I^{\mathrm{used}}_{im}\subseteq I_i$; paired identifiers do not guarantee identical visual access. The downstream task is
\begin{equation}
\widehat I_{im}=G(Q,y_{im}),\qquad
\widehat c_{im}=R(\widehat I_{im}),
\label{eq:reuse}
\end{equation}
where $Q$ is a common wrapper and $R$ is a fixed culture recogniser. Neither receives source images, additional metadata, or reference culture $C_i$. A match measures pipeline-level cultural-attribution recoverability, jointly affected by description content, generation and recognition.

\subsection{Dataset}
\label{sec:data}
We use the artefact corpus released with~\cite{jiang2026appear2meaningcrossculturalbenchmarkstructured}, which provides source images and catalogue metadata from The Metropolitan Museum of Art (Met)\footnote{\url{https://www.metmuseum.org/art/collection/}. Public-domain images and basic collection data are available under CC0.} and the J. Paul Getty Museum (Getty)\footnote{\url{https://www.getty.edu/art/collection/}. Public-domain Open Content images are under CC0; collection-record text and descriptive metadata are under CC BY 4.0 unless otherwise noted, subject to object-level rights statements.}
We evaluate six open-weight and nine closed-source description models on the same 750 museum artefacts, with 50 artefacts in each of fifteen populated region-type strata ($15\times750=11{,}250$ model-artefact records). Table~\ref{tab:description-models} lists the models. Every artefact joins to a nonempty object-specific culture reference from this corpus; the complete records and their qualified or alternative labels remain unchanged. Regions define sampling strata, not culture labels.

\begin{table}[!b]
\centering\small
\setlength{\tabcolsep}{2.6pt}
\caption{Description models and final language-eligibility rates over 750 original output slots per model. Year denotes the initial release of the named model version or family. Bold models exceed 90\% and enter downstream evaluation; the six open-weight models are deployed locally.}
\label{tab:description-models}
\label{tab:language-pass}
\label{tab:cohort-models}
\begin{tabularx}{\columnwidth}{@{}Ylrr@{}}
\toprule
Model & Org. & Year & Pass (\%) \\
\midrule
\multicolumn{4}{c}{\textbf{Open-weight models}} \\
Qwen3-VL-32B~\cite{qwen3_vl} & Alibaba & 2025 & 33.60 \\
\textbf{LLaVA-OneVision-1.5-8B}~\cite{an2025llavaonevision15} & LMMs-Lab & 2025 & \textbf{95.73} \\
DeepSeek-VL2~\cite{wu2024deepseekvl2} & DeepSeek & 2024 & 5.33 \\
Idefics2-8B~\cite{huggingface2024idefics2} & Hugging Face & 2024 & 88.40 \\
Llama-3.2-11B-Vision-Instruct~\cite{meta2024llama32} & Meta & 2024 & 60.40 \\
Pixtral-12B~\cite{mistral2024pixtral12b} & Mistral AI & 2024 & 78.27 \\
\midrule
\multicolumn{4}{c}{\textbf{Closed-source models}} \\
\textbf{GPT-4.1-mini}~\cite{openai_gpt41} & OpenAI & 2025 & \textbf{93.73} \\
GPT-5.4~\cite{openai2026gpt54release} & OpenAI & 2026 & 73.73 \\
GPT-5.4-mini~\cite{openai2026gpt54mini} & OpenAI & 2026 & 79.07 \\
\textbf{Gemini-2.5-Flash}~\cite{google_gemini25} & Google & 2025 & \textbf{90.53} \\
\textbf{Gemini-2.5-Flash-Lite}~\cite{google_gemini25} & Google & 2025 & \textbf{93.73} \\
\textbf{Gemini-3-Flash-Preview}~\cite{google2025gemini3flash} & Google & 2025 & \textbf{94.40} \\
Claude Haiku 4.5~\cite{anthropic2025haiku45} & Anthropic & 2025 & 88.13 \\
Qwen3-VL-Plus~\cite{alibaba2026model_lifecycle} & Alibaba & 2025 & 52.27 \\
Qwen3-VL-Flash~\cite{alibaba2026model_lifecycle} & Alibaba & 2025 & 75.33 \\
\bottomrule
\end{tabularx}
\end{table}

\paragraph{Language eligibility.}
A fixed text-only screen combines CoLA grammatical acceptability~\cite{warstadt2019acceptability} with GPT-4.1-mini judgements of coherence and textual completeness. A description must pass all criteria; factual or cultural correctness is outside this gate, and the original text is never rewritten. For eligibility indicator $e_{im}$,
\begin{equation}
\begin{aligned}
q_m&=\frac{1}{750}\sum_{i=1}^{750}e_{im},\qquad
\mathcal M_5=\{m:q_m>0.90\},\\
\mathcal U&=\bigcap_{m\in\mathcal M_5}\{i:e_{im}=1\},\qquad
|\mathcal U|=570.
\end{aligned}
\label{eq:shared-cohort}
\end{equation}
The denominator includes all 750 original output slots per model. Table~\ref{tab:description-models} reports final eligibility rates and identifies the five retained models.

\paragraph{Eligibility filtering and sampling.}
To reduce image-generation failures, we screen original metadata for potentially sensitive vocabulary, including nudity- and violence-associated terms, while retaining historical East Asian arms and armour catalogued as metalwork. Metadata and generation-feasibility filtering leave 521 eligible artefacts (Table~\ref{tab:cohort-strata}); the refusal and replacement rule is specified in Appendix~\ref{appendix:limitation}.

\begin{table}[!b]
\centering\small
\setlength{\tabcolsep}{3.4pt}
\caption{Eligible artefacts after metadata and generation-feasibility filtering. East Asia--sculpture has no original stratum. The selected quota is 25 per stratum except Europe--sculpture (24); sampling is without replacement.}
\label{tab:cohort-strata}
\begin{tabular}{@{}lrrrrr@{}}
\toprule
Region & Cer. & Metal & Paint & Sculp. & Total \\
\midrule
Americas & 41 & 31 & 39 & 30 & 141 \\
Ancient Med. & 29 & 42 & 32 & 34 & 137 \\
East Asia & 46 & 36 & 42 & --- & 124 \\
Europe & 35 & 35 & 25 & 24 & 119 \\
\midrule
Total & 151 & 144 & 138 & 88 & 521 \\
\bottomrule
\end{tabular}
\end{table}

From the resulting pool, the released seeded procedure selects 374 distinct artefacts with per-stratum sizes $n_h$:
\begin{equation}
\begin{aligned}
\mathcal S&=\bigcup_{h=1}^{15}\mathcal S_h,\quad |\mathcal S|=374,\\
|\mathcal S_h|=n_h&=\begin{cases}24,&h=\text{Europe--sculpture},\\25,&\text{otherwise}.\end{cases}
\end{aligned}
\label{eq:stratified-draw}
\end{equation}
The same selected artefacts are paired across all five models. Their complete reference records are evaluator-only; no reference fields or original images are added to generation prompts.

\subsection{Evaluation Protocol}
\label{sec:evaluation}
\paragraph{Evaluation targets and human baseline.}
The paired target set contains one unchanged description per selected artefact and model. A subset $\mathcal H\subset\mathcal S$ additionally has original museum-authored descriptions, providing a human-description baseline:
\begin{equation}
\begin{aligned}
N_{\mathrm{LLM}}&=5\sum_{h=1}^{15}n_h=1{,}870,\\
N_{\mathrm{human}}&=|\mathcal H|=230,\qquad
N_{\mathrm{total}}=2{,}100.
\end{aligned}
\label{eq:generation-list}
\end{equation}
For both conditions, a fixed generator $G$ receives only the unedited description and common wrapper $Q$ (Equation~\ref{eq:reuse}). The human baseline places museum-authored and model descriptions of the same 230 artefacts in $\mathcal H$ under matched generation, recognition and scoring settings. Museum texts are written from records of the objects for documentation and interpretation, while model outputs follow a common description prompt, so the baseline compares a human describer with model describers on the same objects. The five-model comparison uses all 374 artefacts.

\paragraph{Image generation.}
We reconstruct each unchanged description using OpenAI's native image-generation service through Codex with the \texttt{gpt-6.1-sol} language model, with the same rendering instruction across sources. The generator receives text only and produces one image per condition; execution settings are detailed in Appendix~\ref{appendix:prompt}.

\paragraph{Image-only culture recognition.}
The recogniser is GPT-4.1-mini, selected from the published five-field outputs of~\cite{jiang2026appear2meaningcrossculturalbenchmarkstructured} by Equation~\ref{eq:recognizer-selection}: complete output coverage on the 374-object cohort takes priority, followed by culture agreement. It receives generated image pixels only, without description, title, catalogue metadata or candidate labels, and returns a culture prediction $\widehat c_{im}$ (prompt in Appendix~\ref{appendix:prompt}). Agreement between $\widehat c_{im}$ and the object's reference culture labels $C_i$ is decided by the LLM judge of~\cite{jiang2026appear2meaningcrossculturalbenchmarkstructured}, DeepSeek-Flash with its judging prompt unchanged, which that work validated against human annotation. The same judge scores culture predictions from model-description reconstructions, the human baseline and the original-image control, so all conditions share one agreement criterion. Unknown, refused and unparseable predictions count as non-matches.

\paragraph{Accuracy and uncertainty.}
For cohort $D\in\{\mathcal S,\mathcal H\}$, let $g_{im}$ indicate an available image, $r_{im}$ a valid culture prediction, and $\ell(\widehat c_{im},C_i)$ agreement with the reference. We report fixed-denominator culture accuracy:
\begin{equation}
\begin{aligned}
z_{im}&=g_{im}r_{im}\ell(\widehat c_{im},C_i),\\
\operatorname{Acc}_m(D)&=\frac{1}{|D|}\sum_{i\in D}z_{im}.
\end{aligned}
\label{eq:paired-recovery}
\end{equation}
Generation refusals and unknown or invalid predictions receive zero. Comparisons use $|\mathcal S|=374$ or the matched $|\mathcal H|=230$. Primary accuracy and paired net-change 95\% intervals use 20,000 stratified artefact bootstrap resamples, retaining each object's original and reconstructed outcomes together. Appendix Table~\ref{tab:bootstrap-auxiliary} reports the intervals, coverage, valid-output accuracy and equal-stratum macro accuracy.

\paragraph{Control evaluation.}
The control applies the same recogniser and culture prompt to source photographs, bypassing description and generation. It measures what the recogniser reads from the object itself, so each reconstruction can be compared with its own source as retained, lost or gained agreement.

\newcommand{\MainCultureResultsTable}{%
\begin{table}[t]
\centering\small
\setlength{\tabcolsep}{4pt}
\caption{Culture accuracy (\%) after reconstruction. The human baseline uses the same generator, recogniser and scoring as model descriptions and is available on $\mathcal H$ only. The control applies the same recogniser to original photographs ($195/374$, $122/230$). Accuracy uses fixed denominators (Equation~\ref{eq:paired-recovery}); 95\% intervals are in Appendix Table~\ref{tab:bootstrap-auxiliary}.}
\label{tab:main}
\begin{tabularx}{\columnwidth}{@{}Yrr@{}}
\toprule
Condition & $\mathcal S$ ($N{=}374$) & $\mathcal H$ ($N{=}230$) \\
\midrule
\multicolumn{3}{@{}l}{\textit{Model description $\rightarrow$ generated image}} \\
LLaVA-OneVision-1.5-8B & 54.81 & 52.61 \\
Gemini-3-Flash-Preview & 54.55 & 54.78 \\
GPT-4.1-mini & 59.09 & 57.83 \\
Gemini-2.5-Flash-Lite & 57.75 & 57.83 \\
Gemini-2.5-Flash & 55.88 & 54.35 \\
\midrule
\multicolumn{3}{@{}l}{\textit{Human baseline: museum description $\rightarrow$ generated image}} \\
Museum-authored & n/a & 52.17 \\
\midrule
\multicolumn{3}{@{}l}{\textit{Control: original photograph, no description}} \\
Recogniser on source images & 52.14 & 53.04 \\
\bottomrule
\end{tabularx}
\end{table}%
}

\newcommand{\PairedChangeResultsTable}{%
\begin{table}[t]
\centering\footnotesize
\setlength{\tabcolsep}{3pt}
\caption{Paired outcomes relative to original photographs (\%). Kept: both correct; Lost: original only; Gained: reconstruction only; Neither: neither correct. Net $=$ Gained $-$ Lost. (a) Five models on $\mathcal S$ ($374$); (b) all sources on the same $\mathcal H$ ($230$); (c) regions on $\mathcal S$, pooled over five models.}
\label{tab:paired}
\begin{tabularx}{\columnwidth}{@{}Yrrrrr@{}}
\toprule
 & Kept & Lost & Gained & Neither & Net \\
\midrule
\multicolumn{6}{@{}l}{\textit{(a) Full cohort $\mathcal S$ ($N=374$)}} \\
LLaVA-OneVision-1.5-8B & 42.78 & 9.36 & 12.03 & 35.83 & $+2.67$ \\
Gemini-3-Flash-Preview & 45.19 & 6.95 & 9.36 & 38.50 & $+2.41$ \\
GPT-4.1-mini & 45.45 & 6.68 & 13.64 & 34.22 & $+6.95$ \\
Gemini-2.5-Flash-Lite & 44.39 & 7.75 & 13.37 & 34.49 & $+5.61$ \\
Gemini-2.5-Flash & 45.19 & 6.95 & 10.70 & 37.17 & $+3.74$ \\
\midrule
\multicolumn{6}{@{}l}{\textit{(b) Matched cohort $\mathcal H$ ($N=230$)}} \\
LLaVA-OneVision-1.5-8B & 41.30 & 11.74 & 11.30 & 35.65 & $-0.43$ \\
Gemini-3-Flash-Preview & 43.91 & 9.13 & 10.87 & 36.09 & $+1.74$ \\
GPT-4.1-mini & 42.61 & 10.43 & 15.22 & 31.74 & $+4.78$ \\
Gemini-2.5-Flash-Lite & 43.48 & 9.57 & 14.35 & 32.61 & $+4.78$ \\
Gemini-2.5-Flash & 43.04 & 10.00 & 11.30 & 35.65 & $+1.30$ \\
Museum-authored & 41.30 & 11.74 & 10.87 & 36.09 & $-0.87$ \\
\midrule
\multicolumn{6}{@{}l}{\textit{(c) By region, pooled over the five models}} \\
Americas & 20.60 & 4.40 & 12.20 & 62.80 & $+7.80$ \\
Ancient Mediterranean & 47.80 & 8.20 & 13.40 & 30.60 & $+5.20$ \\
East Asia & 78.40 & 1.60 & 8.00 & 12.00 & $+6.40$ \\
Europe & 40.00 & 14.55 & 12.73 & 32.73 & $-1.82$ \\
\bottomrule
\end{tabularx}
\end{table}%
}

\newcommand{\ObjectIdentificationResultsTable}{%
\begin{table}[t]
\centering\small
\setlength{\tabcolsep}{3.5pt}
\caption{Closed-book object identification (\%). Text uses \texttt{gpt-6.1-sol}; images use GPT-4.1-mini. Claimed: identity asserted; Verified: catalogue-confirmed. Claimed/Verified denominators are all supplied inputs: model text $374$ each, museum text $230$; original images $374/230$; generated images $1{,}821/1{,}338$ (five/six sources). Text Acc.$_{\neg\mathrm{ID}}$ uses $\mathcal S_0$ ($300$) or $\mathcal H_0$ ($145$); image-conditioned denominators are in Appendix~\ref{appendix:metric-definitions}.}
\label{tab:leakage}
\begin{tabularx}{\columnwidth}{@{}Yrrr@{}}
\toprule
Input & Claimed & Verified & Acc.$_{\neg\mathrm{ID}}$ \\
\midrule
\multicolumn{4}{@{}l}{\textit{(a) Text side: description $\rightarrow$ \texttt{gpt-6.1-sol}}} \\
LLaVA-OneVision-1.5-8B & 10.96 & 7.49 & 56.33 \\
Gemini-3-Flash-Preview & 11.23 & 5.35 & 56.00 \\
GPT-4.1-mini & 12.30 & 7.75 & 60.33 \\
Gemini-2.5-Flash-Lite & 7.49 & 5.35 & 57.33 \\
Gemini-2.5-Flash & 11.76 & 7.22 & 57.33 \\
Museum-authored & 32.61 & 9.57 & 51.72 \\
\midrule
\multicolumn{4}{@{}l}{\textit{(b) Image side: image $\rightarrow$ recogniser}} \\
Original photographs, $\mathcal S$ & 2.67 & 0.80 & 53.58 \\
Original photographs, $\mathcal H$ & 1.74 & 0.00 & 49.31 \\
Generated images, $\mathcal S$ & 0.49 & 0.05 & 57.37 \\
Generated images, $\mathcal H$ & 0.97 & 0.00 & 54.10 \\
\bottomrule
\end{tabularx}
\end{table}%
}

\section{Experiment}
\label{sec:experiment}
We ask what model descriptions keep of each object by comparing their reconstructions with the source photographs, with museum descriptions of the same objects, and with identity probes.

\subsection{Recogniser Selection}
\label{sec:control}
\begin{table}[t]
\centering\small
\setlength{\tabcolsep}{4pt}
\caption{Culture agreement recomputed from the five-field model outputs published with~\cite{jiang2026appear2meaningcrossculturalbenchmarkstructured} on $\mathcal S$ ($N{=}374$), for recogniser selection. Among complete-coverage endpoints, GPT-4.1-mini has the highest culture accuracy and is selected as $r^\star$. All nine endpoints are in Appendix Table~\ref{tab:control-current}.}
\label{tab:recognizer}
\begin{tabularx}{\columnwidth}{@{}Yrr@{}}
\toprule
Endpoint $r\in\mathcal R_{374}$ & $A_r^{\mathrm{orig}}$ & $100A_r^{\mathrm{orig}}/N$ \\
\midrule
\textbf{GPT-4.1-mini} & \textbf{216} & \textbf{57.75} \\
GPT-5.4-mini & 213 & 56.95 \\
Qwen3-VL-32B & 177 & 47.33 \\
Qwen3-VL-8B & 164 & 43.85 \\
\bottomrule
\end{tabularx}
\end{table}
For each published endpoint $r$ on a cohort of $N$ objects, $V_r^{\mathrm{orig}}$ counts valid original-image outputs and $A_r^{\mathrm{orig}}$ counts culture matches. We select
\begin{equation}
\mathcal R_N=\{r: V_r^{\mathrm{orig}}=N\},\quad
r^\star=\operatorname*{arg\,max}_{r\in\mathcal R_N}\frac{A_r^{\mathrm{orig}}}{N},
\label{eq:recognizer-selection}
\end{equation}
and set $R=f_{r^\star}$. In the published five-field outputs, four of nine endpoints have complete coverage; GPT-4.1-mini is highest at $216/374=57.75\%$ (Table~\ref{tab:recognizer}). These archived five-field scores select the endpoint; the subsequent culture-only control supplies the paired reference for reconstruction. 

\subsection{Main Results}
\label{sec:main}
\paragraph{Image-generation execution.}
Of 2,100 description conditions for the 374-object cohort, 2,051 produce images ($97.67\%$) and 49 receive safety refusals ($2.33\%$; Table~\ref{tab:generation-progress}). All descriptions remain unchanged.

\begin{table}[t]
\centering\footnotesize
\setlength{\tabcolsep}{2pt}
\caption{Image execution for the full stratified cohort (374 retained artefacts, 25 per stratum except Europe--sculpture (24)). Counts concern image availability, not culture accuracy; Success $=$ Images$/$Conditions (\%).}
\label{tab:generation-progress}
\begin{tabularx}{\columnwidth}{@{}Yrrrr@{}}
\toprule
Description source & Conditions & Images & Refusals & Success \\
\midrule
\mbox{LLaVA-OneVision-1.5-8B} & 374 & 362 & 12 & 96.79 \\
Gemini-3-Flash-Preview & 374 & 366 & 8 & 97.86 \\
GPT-4.1-mini & 374 & 365 & 9 & 97.59 \\
Gemini-2.5-Flash-Lite & 374 & 364 & 10 & 97.33 \\
Gemini-2.5-Flash & 374 & 364 & 10 & 97.33 \\
Museum-authored & 230 & 230 & 0 & 100.00 \\
\midrule
Total & 2,100 & 2,051 & 49 & 97.67 \\
\bottomrule
\end{tabularx}
\end{table}

\MainCultureResultsTable
\PairedChangeResultsTable
\ObjectIdentificationResultsTable

\begin{figure}[t]
\centering
\includegraphics[width=\columnwidth]{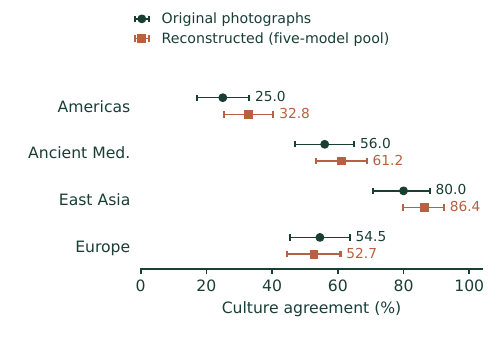}
\caption{Regional culture recovery on $\mathcal S$: original photographs and reconstructions pooled over five models. Whiskers are 95\% stratified artefact bootstrap intervals. Regional object counts are 100, 100, 75 and 99, respectively.}
\label{fig:regional-recovery}
\end{figure}

Table~\ref{tab:main} compares the full cohort $\mathcal S$ and the matched museum-description subset $\mathcal H$. Table~\ref{tab:paired} cross-tabulates each object's original and reconstructed outcomes; every source in panel (b) uses the same 230 objects.

\paragraph{Analysis.}
\emph{(i) Regional disparities dominate recovery.} East Asia and the Americas score $80.0\%$ versus $25.0\%$ on original photographs and $86.4\%$ versus $32.8\%$ after reconstruction, pooled over the five models (Figure~\ref{fig:regional-recovery}). Their gap remains $53.6$ points after reuse, against $55.0$ before, far exceeding the $4.55$-point spread between description models. Descriptions carry this unevenness forward rather than correct it. Appendix Table~\ref{tab:regional-uncertainty} gives regional intervals.
\emph{(ii) Small aggregate gains hide many object-level changes.} On $\mathcal S$, point changes $\Delta_m=\operatorname{Acc}_m-\operatorname{Acc}_{\mathrm{ctrl}}$ range from $+2.41$ to $+6.95$ points, while $16$--$21\%$ of objects change outcome (Table~\ref{tab:paired}a). GPT-4.1-mini gains $6.95$ points (95\% CI $[2.67,11.50]$) and Gemini-2.5-Flash-Lite $5.61$ ($[1.34,10.16]$); the other three paired intervals include zero.
\emph{(iii) Models add more type than curators.} On $\mathcal H$, museum-authored text loses $11.74\%$ and gains $10.87\%$. The five models lose $9.13$--$11.74\%$ and gain $10.87$--$15.22\%$ on those same objects (Table~\ref{tab:paired}b). Museum text and LLaVA share the highest loss rate; their net changes are $-0.87$ and $-0.43$ points. Models thus lose recorded cultures at a rate similar to curators but gain more, moving objects toward what their culture typically looks like.

\begin{figure*}[!t]
\centering
\includegraphics[width=\textwidth]{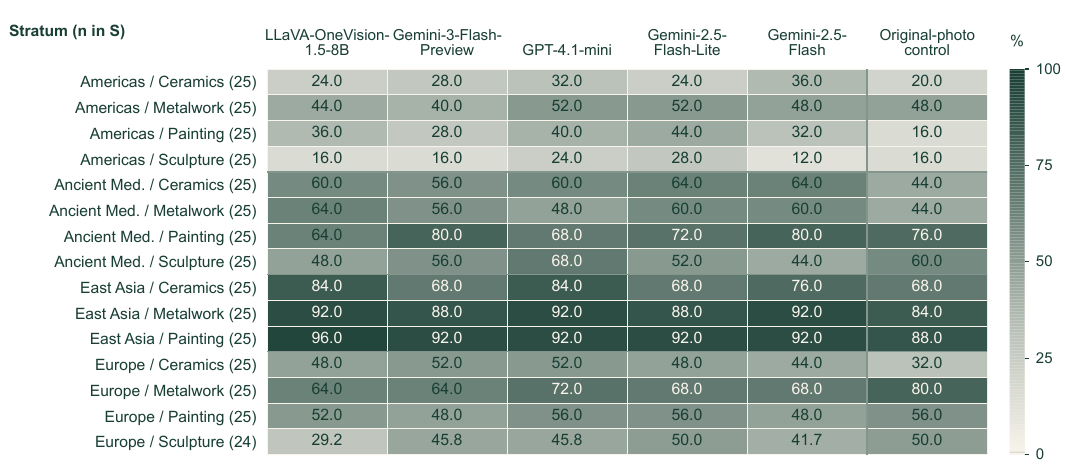}
\caption{Descriptive culture agreement (\%) by region $\times$ type on $\mathcal S$ ($N=374$): five description models and the original-photograph control. Fixed denominators are 25 per stratum, except Europe--sculpture (24); refusals and unknown predictions count as non-matches.}
\label{fig:stratum-recovery}
\end{figure*}

\begin{table*}[!t]
\centering
\small
\begingroup
\setlength{\tabcolsep}{4pt}
\renewcommand{\arraystretch}{1.08}
\caption{Object-specific discrepancies in seven qualitative cases, with original-image thumbnails, recorded references and verbatim candidate excerpts. Case 6 quotes reported fragments.}
\label{tab:cases}
\begin{tabular}{@{}
  >{\centering\arraybackslash}m{0.085\textwidth}
  >{\raggedright\arraybackslash}m{0.26\textwidth}
  >{\raggedright\arraybackslash}m{0.30\textwidth}
  >{\raggedright\arraybackslash}m{\dimexpr0.355\textwidth-6\tabcolsep\relax}
  @{}}
\toprule
\multicolumn{2}{@{}l}{\textbf{Case and recorded reference}} &
\textbf{Representative candidate excerpt} &
\textbf{Diagnostic observation} \\
\midrule
\includegraphics[width=1.25cm,height=1.45cm,keepaspectratio]{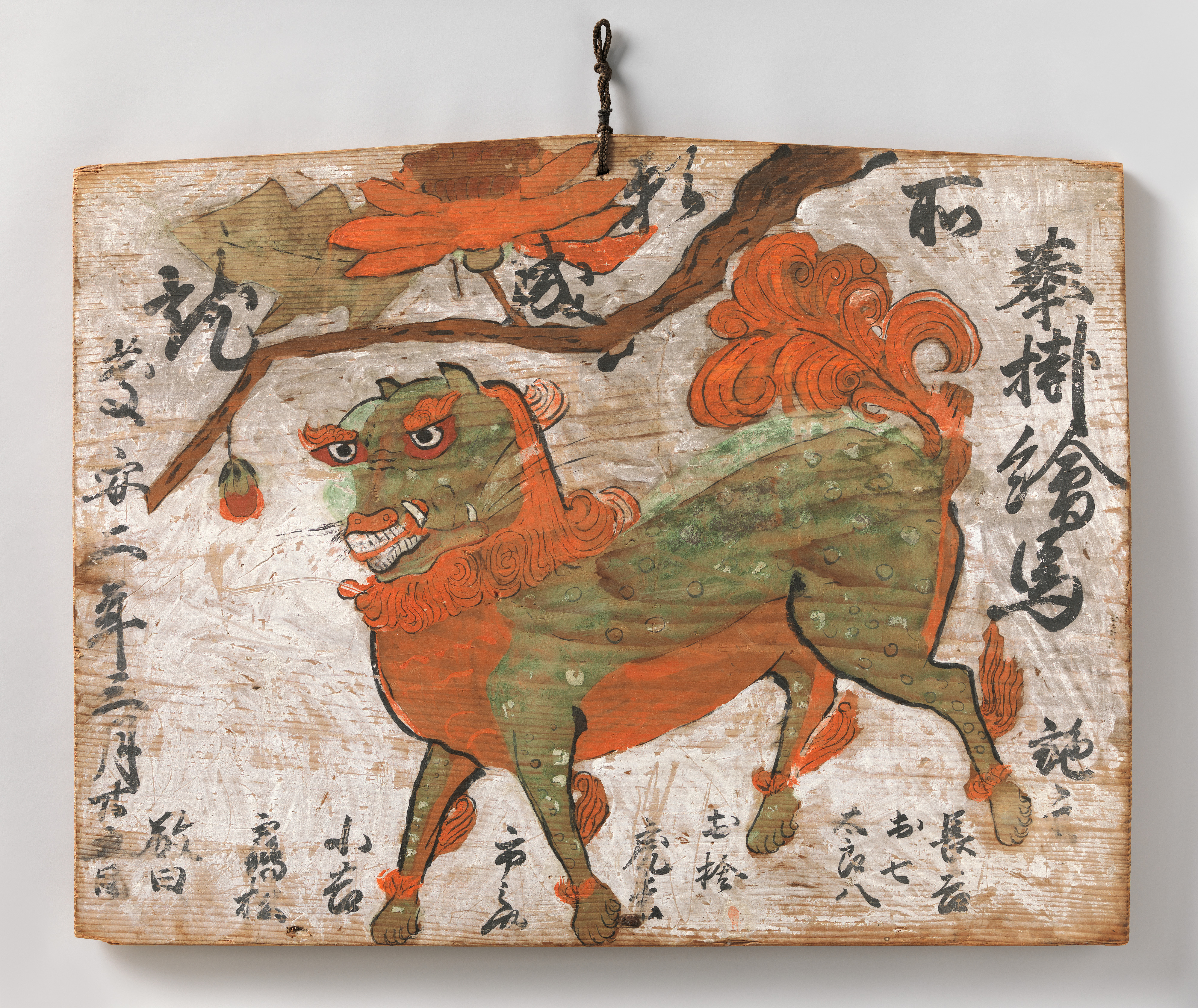} &
\textbf{1. Votive lion.} Japanese votive painting, 1649; lion, peony and inscriptions. &
``protective creature in East Asian tradition''; ``syncretic religious practice of votive offerings at Japanese shrines'' &
General symbolism gives less emphasis to the inscription--image relation than reader notes; the precise inscription reading remains unverified. \\
\addlinespace[5pt]
\includegraphics[width=1.25cm,height=1.45cm,keepaspectratio]{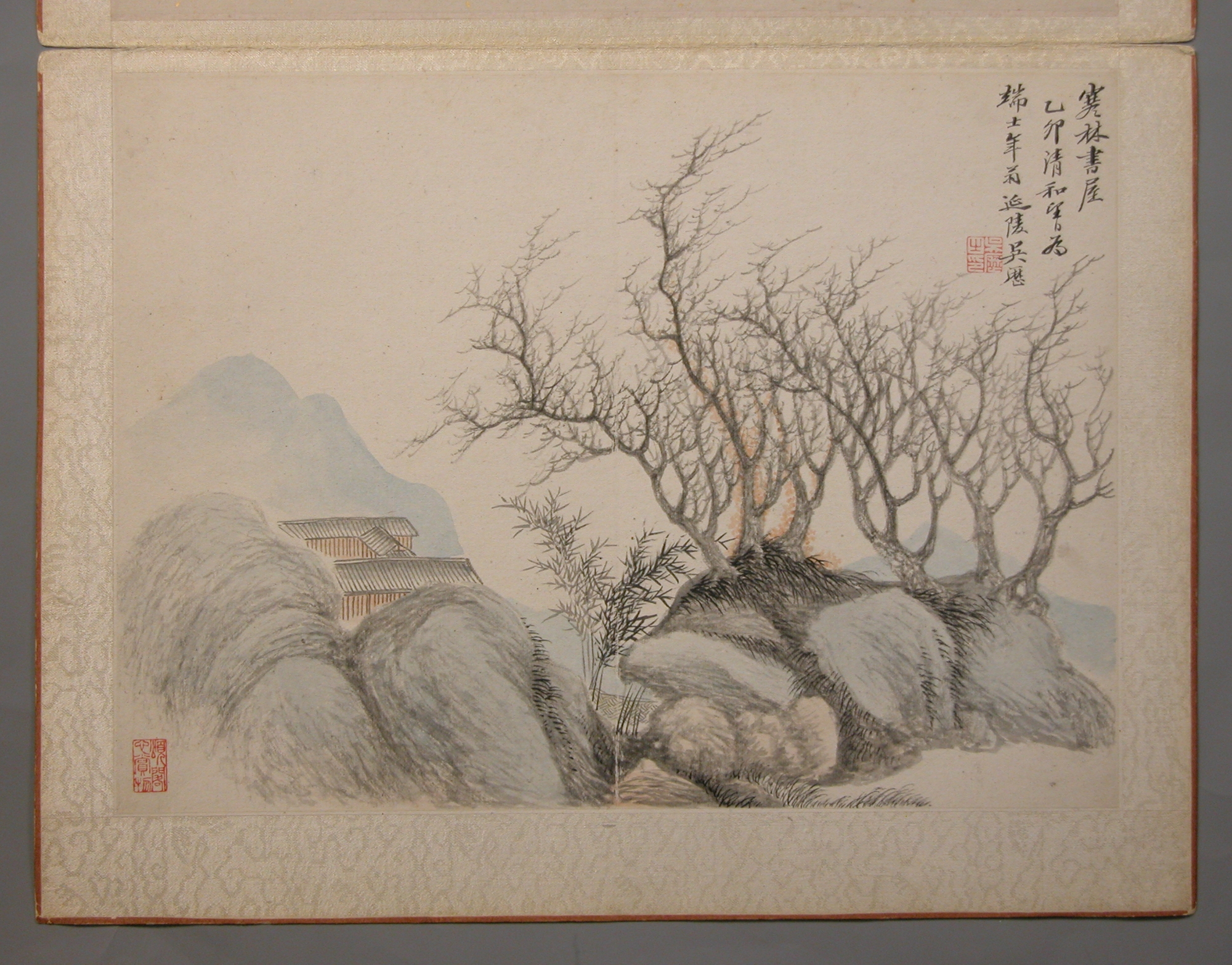} &
\textbf{2. Landscape album.} Chinese; Wang Jian (supplied attribution), 1674--76; foreground trees and rocks. &
``literati painting tradition''; ``historical precedents and regional landscape traditions'' &
Broad art-historical framing gives less prominence to foreground trees and rocks than the readers' account. \\
\addlinespace[5pt]
\includegraphics[width=1.25cm,height=1.45cm,keepaspectratio]{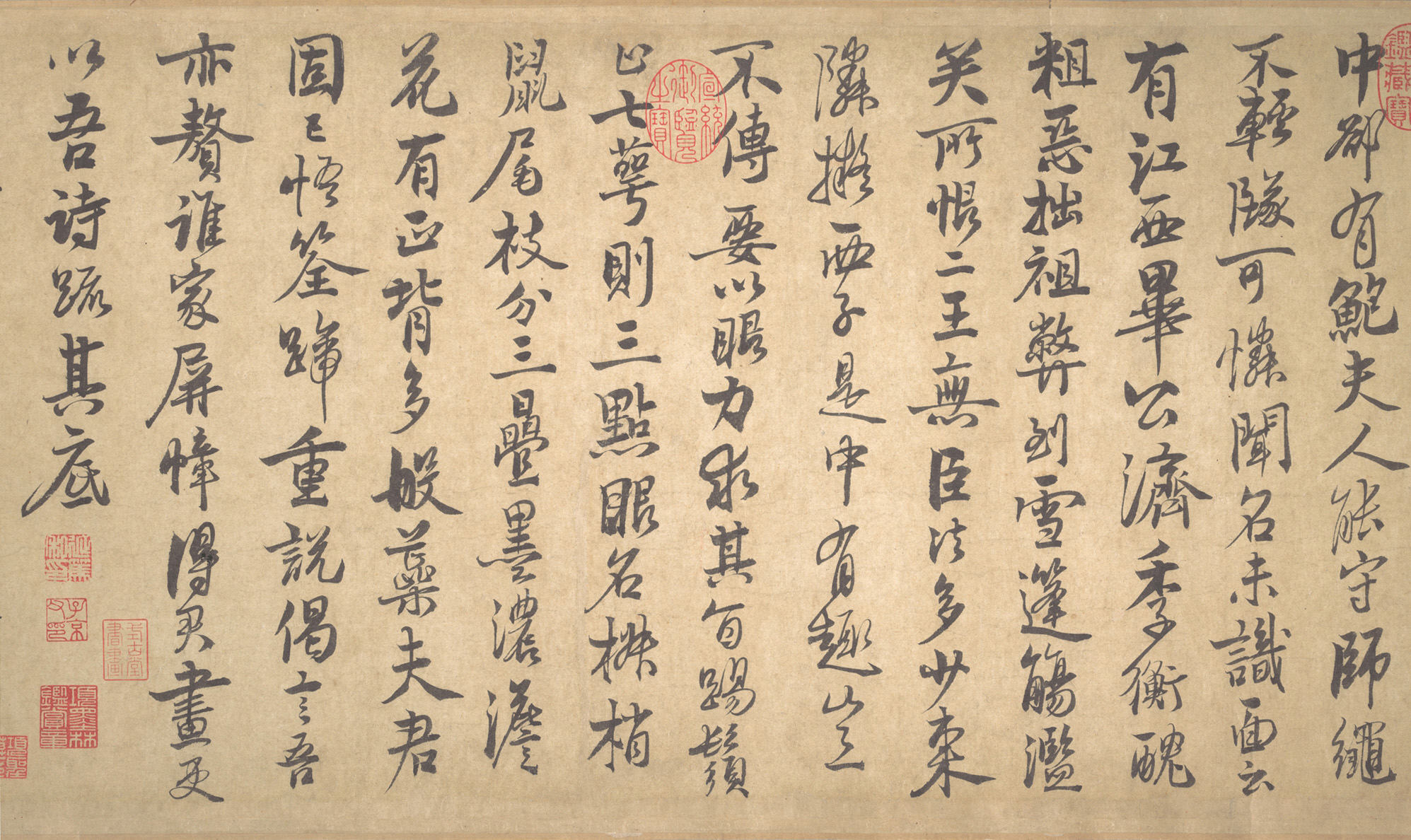} &
\textbf{3. Calligraphic poems.} Title: \emph{Poems on painting plum blossoms and bamboo}; writing and seals. &
``celebrating the subjects of plum blossoms and bamboo through calligraphy''; ``moral character and philosophical virtue'' &
Recognises calligraphy but elaborates title-linked plant symbolism; readers foreground writing and seals. A causal title effect is untested. \\
\addlinespace[5pt]
\includegraphics[width=1.25cm,height=1.45cm,keepaspectratio]{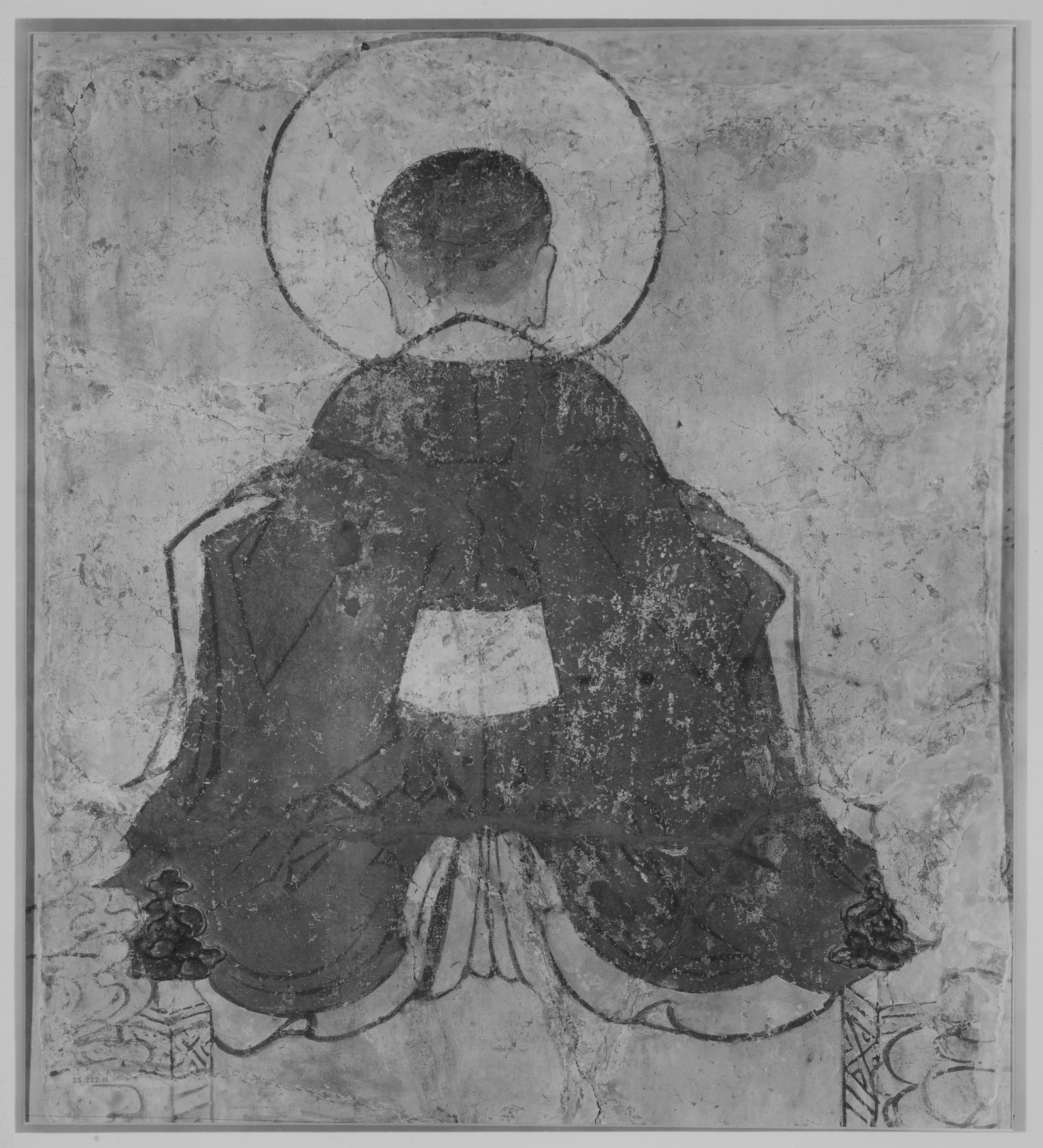} &
\textbf{4. Wall painting.} Chinese; readers interpret a back-facing figure with a railing in front. &
``The figure is shown in frontal pose''; ``religious, ancestral, or imperial figure'' &
Frontal pose conflicts with the readers' interpretation; generic religious framing leaves object-specific orientation and identity unresolved. \\
\addlinespace[5pt]
\includegraphics[width=1.25cm,height=1.45cm,keepaspectratio]{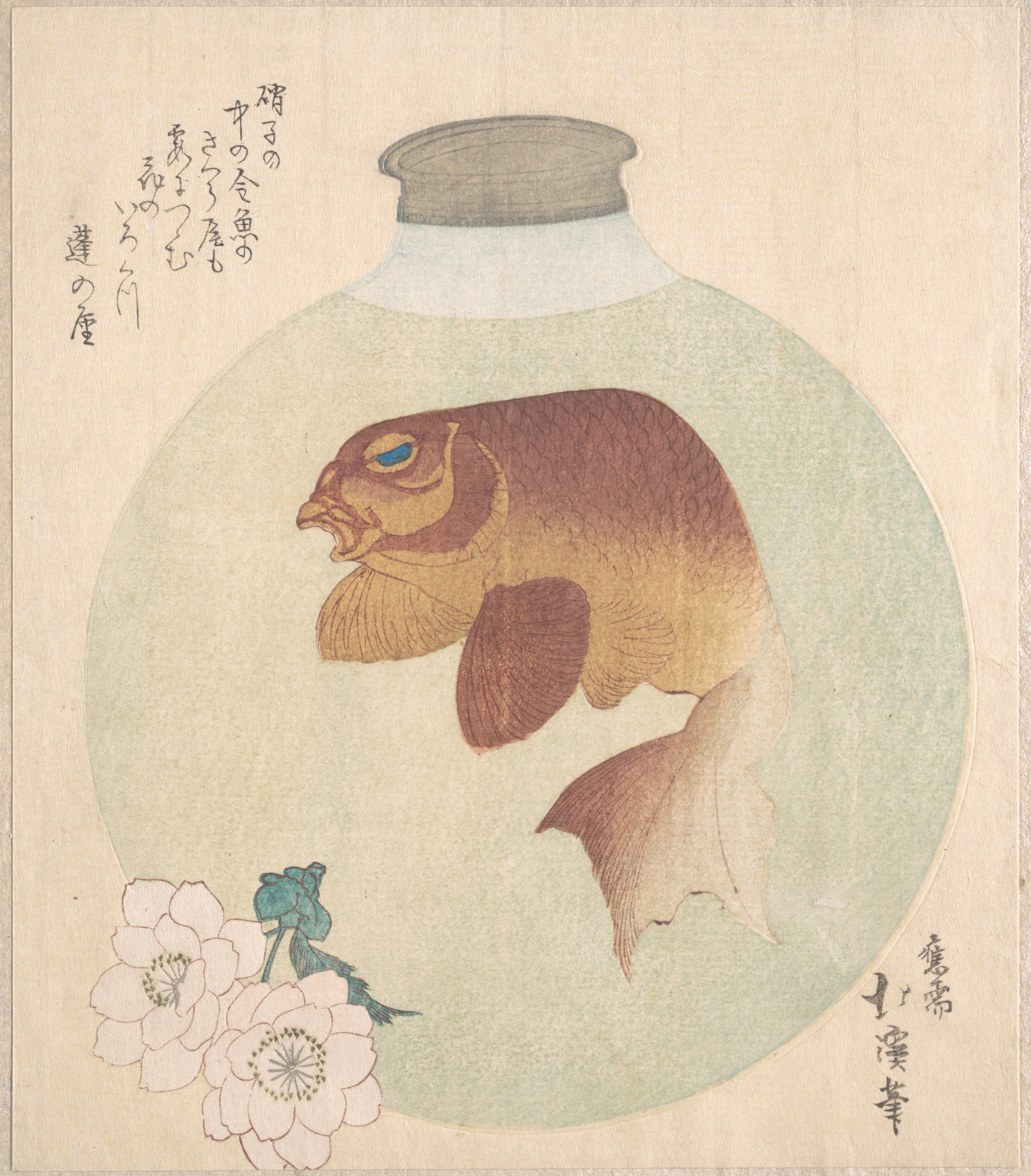} &
\textbf{5. Goldfish print.} Japanese woodblock print; goldfish in a glass bottle. &
``may suggest the influence of Western representational techniques on Japanese printmaking'' &
The qualified influence hypothesis lacks object-specific support in the available notes; this does not establish that it is false. \\
\addlinespace[5pt]
\includegraphics[width=1.25cm,height=1.45cm,keepaspectratio]{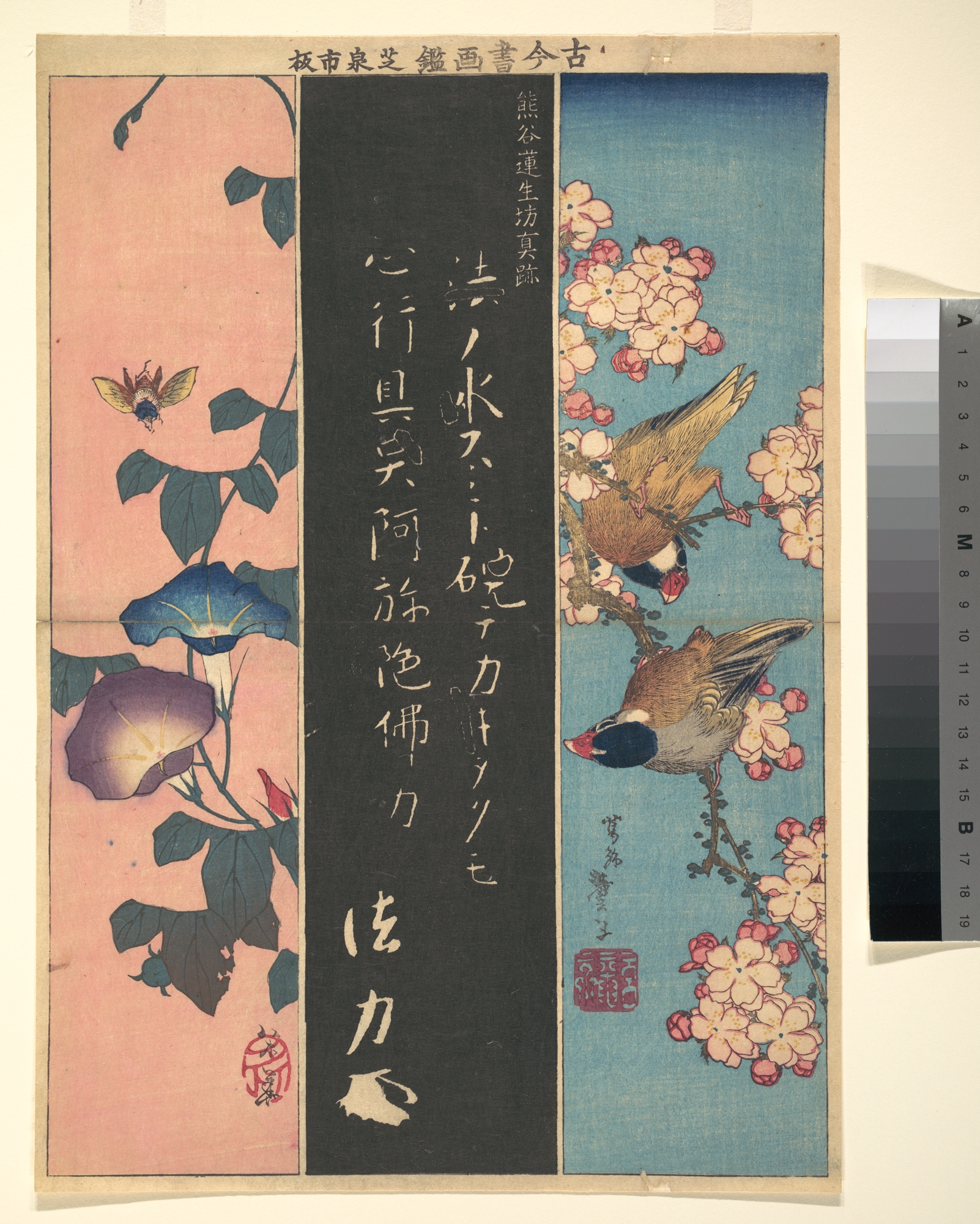} &
\textbf{6. Bird-and-flower print.} Japanese; Hokusai attribution appears in the case notes. &
\emph{Reported excerpts only:} ``Japanese text in gold''; ``small white and coral blossoms'' &
Gold is a material claim beyond visible yellow colour; gold, paper ageing and botanical readings all require documentary support. \\
\addlinespace[5pt]
\includegraphics[width=1.25cm,height=1.45cm,keepaspectratio]{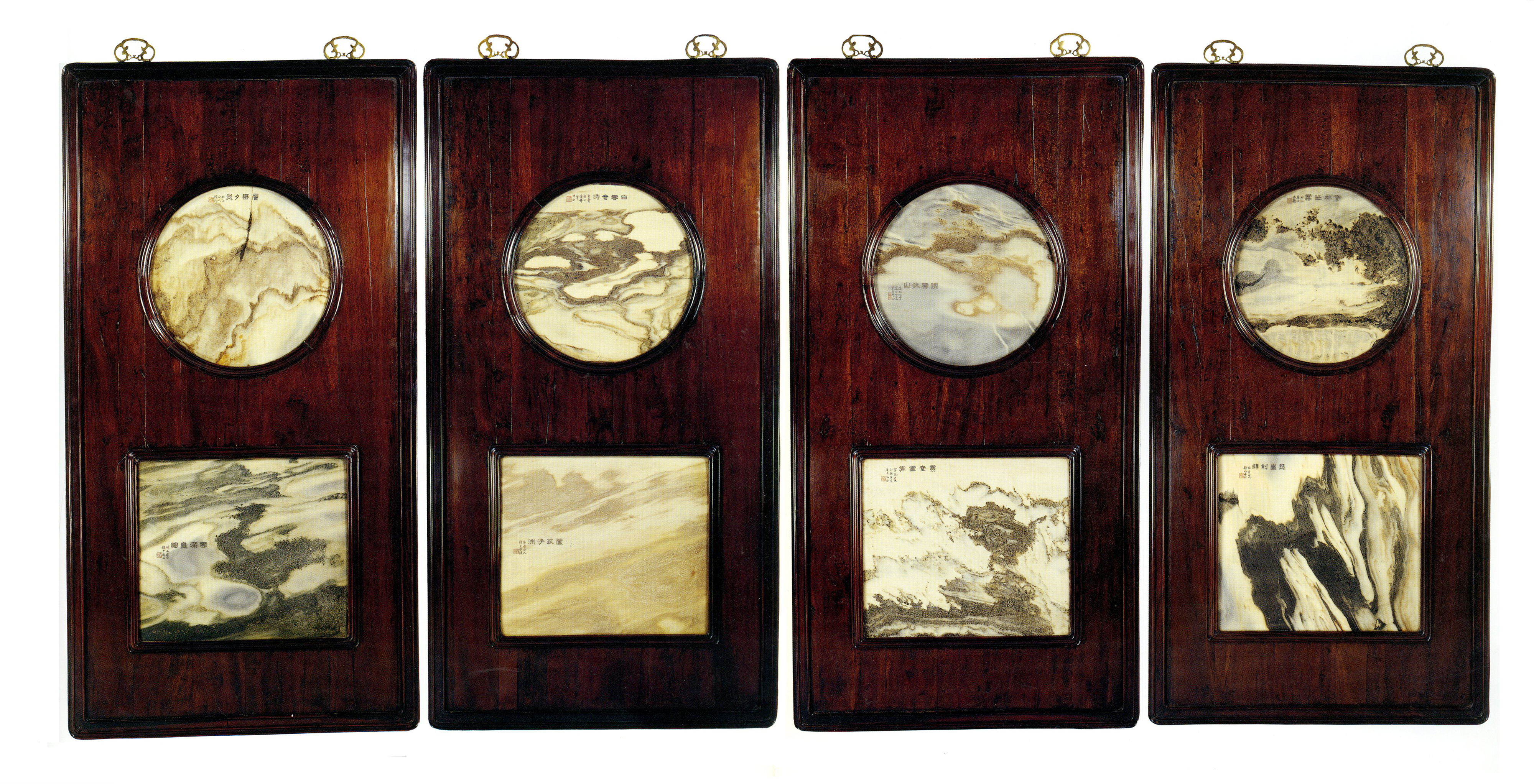} &
\textbf{7. Stone panels.} Chinese; supplied account describes naturally patterned stone, selected and mounted. &
``executed in ink on stone with subtle tonal gradations''; ``literati painting tradition'' &
Where the natural-pattern account is substantiated, ink-on-stone framing mischaracterises making and human selection. \\
\bottomrule
\end{tabular}
\endgroup
\end{table*}

\emph{(iv) Museum descriptions point to particular objects more often.} On the same 230 objects, museum-authored descriptions elicit 75 identity claims ($32.61\%$), compared with $7.83$--$13.04\%$ for model descriptions (Appendix Table~\ref{tab:matched-identity}). Only 22 museum-text claims are catalogue-verified ($9.57\%$ of descriptions; $29.33\%$ of claims); 13 fail local verification and 40 lack sufficient identifying information. The text-side probe therefore often asserts a specific identity without a catalogue-supported match.
\emph{(v) Specific-object recognition from reconstructed images is rare.} The image-side recogniser claims an identity for $0.49\%$ of generated images, compared with $2.67\%$ of originals (Table~\ref{tab:leakage}). Reconstructions keep the culture of an object far more often than the object itself.
\paragraph{Generator-side recognition screening.}
To reduce leakage risk from recognisable collection items, \texttt{gpt-6.1-sol} assesses each unchanged description in an independent, tool-free request. It returns a recognition decision and, when claimed, the object's title, institution and accession number (Table~\ref{tab:leakage}). Let $u_{im}=1$ only for a valid explicit nonrecognition. The common retained sets are
\begin{equation}
\begin{aligned}
\mathcal S_0&=\{i\in\mathcal S:\prod_{m\in\mathcal M_5}u_{im}=1\},\\
\mathcal H_0&=\{i\in\mathcal H\cap\mathcal S_0:u_{iH}=1\}.
\end{aligned}
\label{eq:nonrecognition-cohort}
\end{equation}
Claimed but unverified identities, inconsistent responses and execution failures are excluded. The screen retains $|\mathcal S_0|=300$ and $|\mathcal H_0|=145$. Model recovery is $56.00$--$60.33\%$ on $\mathcal S_0$; on $\mathcal H_0$, models reach $53.10$--$58.62\%$ versus $51.72\%$ for museum-authored descriptions and a $49.66\%$ original-image control. Tables~\ref{tab:main} and~\ref{tab:paired} provide the full-cohort reference. The image-side probe separately records recognition claims from original and generated images; the screen addresses recognisable-object risk under these requests.

\subsection{Error Analysis and Case Study}
\label{sec:errors}
\paragraph{Quantitative analysis.}
The Americas' low recovery is not accompanied by a large reference-label inventory: 99 of its 100 objects are labelled \emph{American}, and one \emph{Mexican}. Figure~\ref{fig:task} links this regional result to a concrete mismatch: two descriptions name American culture, yet one reconstruction is classified as European Rococo. A plausible stylistic classification can conflict with the object's recorded cultural affiliation. In Table~\ref{tab:paired}c, $62.80\%$ of Americas pairs are unrecognised both before and after reconstruction; Europe is the only region with negative net change ($-1.82$ points). Figure~\ref{fig:stratum-recovery} localises the regional disparity: Americas paintings score $16\%$ on photographs and $28$--$44\%$ after reconstruction, while sculpture scores $16\%$ and $12$--$28\%$, respectively. East Asian metalwork scores $84\%$ on photographs and $88$--$92\%$ after reconstruction. These category-level results show that aggregate gains can coexist with persistently low recovery in particular region-type strata.

\paragraph{Qualitative analysis.}
Table~\ref{tab:cases} compares object-specific features with the candidate descriptions (Appendix~\ref{sec:case_study}). Broad cultural framing includes a ``protective creature in East Asian tradition'' (Case~1) and ``literati painting tradition'' (Case~2). The discrepancies concern the individual object: a frontal pose conflicts with the readers' back-facing interpretation (Case~4), visible yellow becomes ``gold'' (Case~6), and an ``ink on stone'' account contrasts with the supplied natural-pattern account (Case~7). In each case the description keeps a recognisable cultural frame while changing the pose, material or making of the particular object.

\paragraph{Why descriptions give many faces.}
The results fit one account of description. A describer that holds a particular object reproduces that object, while a describer that holds only its kind fills each description with a typical member of the kind, and different descriptions draw different members~\cite{huttenlocher1991categories,bartlett1932remembering}. Vision-language models rarely recover the catalogue attribution of the objects they describe~\cite{jiang2026appear2meaningcrossculturalbenchmarkstructured}, which places them in the second case. Their descriptions keep the culture, which is why reconstructions are recognised more often than the photographs; they lose the object, which is why the specific object is almost never identified; and they differ between models, which is why many objects change outcome. Museum descriptions, written from records of the objects, differ on each count. They gain less, and their text points to a particular object in 32.61\% of cases against 7.83\% to 13.04\% for model text. What a description loses therefore shows what its describer lacks~\cite{bozeat2003duck}, which for these models is knowledge of the particular object rather than of its culture.
\section{Conclusion and Future Work}
\label{sec:conclusion}
\PolytroposBrand{} asks what a new, non-human describer keeps of the cultural objects it describes. Turning descriptions back into images shows that vision-language models keep the culture of an object and lose the object itself. Their reconstructions are recognised as the recorded culture more often than the photographs while the specific object is almost never recognised, and on the same objects they add more type than curators do. Museum text, written from records of the objects, points to particular objects far more often, and American work, recorded as American in 99 of 100 objects, is still read through European styles. These descriptions already enter catalogues, search results, and the training data of later models, where model-typical forms can be amplified with each reuse~\cite{bianchi2023easily,luccioni2023stable,shumailov2024ai}, so what they replace is passed on to present and future readers. Future work should measure instance-level fidelity directly, follow descriptions over repeated reuse, and include descriptions written by source communities, so that the objects of a culture, and not only its types, reach those readers.
\clearpage

\section*{Acknowledgements}
Yuechen Jiang is supported by a PhD studentship jointly funded by the Engineering and Physical Sciences Research Council (EPSRC) [grant number EP/Z534912/1] and the Getty Conservation Institute. The University of Manchester provides additional support through a tuition fee waiver.

\section*{LLM Usage Statement}
\label{sec:llm-usage-statement}
OpenAI Codex assisted with drafting evaluation-metric code, manuscript layout and language polishing, result tables, statistical plots and workflow diagrams, and condensing case descriptions into a table with accompanying Error Analysis. Its role was limited to implementation and presentation assistance. Data selection, task design, prompt design, and the analysis and interpretation of experimental results were carried out by the authors, who remain responsible for the content of this paper.

\bibliographystyle{ieeenat_fullname}
\bibliography{main,revision_refs}

\clearpage
\appendix
\raggedbottom

\section{Limitations}
\label{appendix:limitation}
\paragraph{Collection and reference scope.}
The fifteen-stratum, two-collection cohort is conditional on language eligibility and generation feasibility. Museum references contain qualified, variably granular attributions. Retained cases provide object-specific evidence rather than prevalence estimates.

\paragraph{Pipeline dependence.}
Reconstruction combines description content, generator priors and recogniser error; a culture match can survive altered visual details. Recogniser selection uses the evaluated cohort, and GPT-4.1-mini serves multiple pipeline roles. The image backend exposes no versioned identifier or seed. Bootstrap intervals describe between-artefact variation under one fixed generation and recognition execution. Results cover the five language-eligible description models in one image--text--image loop; prototype-like interpretation draws on qualitative cases, and the identification probe measures query-specific recognisability.

\paragraph{Screening.}
CoLA and GPT screening lack independent human validation, and the follow-up covers first-pass candidates only. Input-field availability and recorded visual access vary. Safety rules condition the population on backend behaviour (Appendix~\ref{appendix:generation-feasibility}). The recognition probe reduces leakage risk from recognisable objects; explicit nonrecognition does not establish absence from training data. Common screening changes the evaluated population.

\section{Ethical Considerations}
\label{appendix:ethical-consideration}
\paragraph{Cultural interpretation and attribution.}
Museum descriptions and labels are situated human accounts. The reference record retains source institutions, alternative attributions and qualifiers. Model outputs are presented as generated interpretations, with their provenance preserved. Catalogue agreement is reported at the object level, and regional strata are sampling categories rather than cultural identities assigned to communities.

\paragraph{Collection material and synthetic images.}
Source images and catalogue text retain their object-level rights statements and scholarly attribution. Public-domain Met images and basic collection data are available under CC0. Getty public-domain Open Content images are under CC0; collection-record text and descriptive metadata are under CC BY 4.0 unless otherwise noted. The release retains source links and rights information. Reconstructions are identified as synthetic images and kept distinct from museum photographs.

\paragraph{Screening and responsible reuse.}
Descriptions remain verbatim throughout screening, generation and analysis. Safety refusals, skipped conditions and replacements are recorded alongside successful executions. Metadata filtering includes nudity- and violence-related vocabulary; historical East Asian arms and armour are retained as metalwork following the feasibility check. Language eligibility concerns readability and textual completion. Cultural interpretations and model-generated catalogue claims require contextual review before reuse in collection records.

\paragraph{Annotation and release practice.}
The corpus-wide language screen is automatic. The earlier case commentary and LLM-assisted pilot notes retain separate provenance. Any additional human study should document expertise, author involvement, informed consent and compensation, and release de-identified decisions and rationales. The anonymous submission omits private account details and internal paths while retaining collection provenance and attribution to prior work.

\section{Prompts and Execution Records}
\label{appendix:prompt}
Table~\ref{tab:appendix-implementation} summarises the executed stages. Prompts and schemas are fixed within each stage; raw responses and format retries retain their own records.

\begin{table*}[t]
\centering\small
\setlength{\tabcolsep}{4pt}
\caption{Implementation specifications for screening, reconstruction and recognition. Output limits are token limits. The image generator receives an unchanged description; image-side recognition receives pixels only.}
\label{tab:appendix-implementation}
\begin{tabularx}{\textwidth}{@{}p{0.15\textwidth}p{0.25\textwidth}p{0.23\textwidth}Y@{}}
\toprule
Role & Model or service & Input & Settings and output \\
\midrule
Sentence screening & \texttt{textattack/}\allowbreak\texttt{roberta-base-CoLA} & Candidate sentences & 128-token chunks; acceptability probability \\
Coherence / completeness & \texttt{gpt-4.1-mini-2025-04-14} & Unchanged description & Temperature 0; strict JSON; two ternary decisions \\
Image reconstruction & Native image service through Codex; language model \texttt{gpt-6.1-sol} & Unchanged description and shared rendering wrapper & Square canvas requested; one archived image per success; execution status and output hash retained \\
Culture recognition & \texttt{gpt-4.1-mini-2025-04-14} & One generated or original image & Temperature 0; high detail; 128 tokens; JSON culture string \\
Culture compatibility & \texttt{deepseek-flash} & Culture prediction and reference array & Temperature 0; 1,200 tokens; binary score and reason \\
Object identification & \texttt{gpt-6.1-sol} (text); GPT-4.1-mini (image) & One description or one image & Strict four-field JSON; 4,096 / 512 tokens; local verification \\
\bottomrule
\end{tabularx}
\end{table*}

\subsection{Description generation}
The fifteen description files use images and the available catalogue metadata under the instruction below. The source snapshot retains \texttt{museum\_description\_pipeline.py} at revision \texttt{33f1838}; the evaluation reference arrays are joined separately. The original description, input fields, image-access status and source identifiers are preserved for each object--model record. Typographic bullets and arrows in the source instruction are rendered as ASCII for pdfLaTeX compatibility.

\begin{Verbatim}[fontsize=\scriptsize,breaklines=true,breakanywhere=true,breaksymbolleft={},breaksymbolright={}]
You are a professional museum curator and collections writer.

Task:
Write a short museum-style description for a cultural heritage object using ONLY the provided structured metadata and one or more object images.

Goal:
Produce a concise, historically informed, culturally grounded, and museum-ethical description of the object.

========================
Core Writing Principles
========================
Your description should:

1. Identify the main subject, scene, or form
2. Select only the most meaningful visible features
3. Interpret those features in relation to the object's historical, cultural, material, or functional context
4. Prioritize cultural and historical significance over exhaustive visual listing

Priority order for selection:

1. Cultural or functional significance
2. Iconography or subject meaning
3. Material or technique (if distinctive)
4. Visual form (only if above are weak)

Do NOT prioritize minor visual details over cultural meaning

========================
CRITICAL: Object-Specificity (REQUIRED)
========================
- Avoid generic descriptions that could apply to many objects
- Ensure at least ONE element is specific to this object
  (e.g., scene, structure, motif, composition, or distinctive feature)

- The description should clearly distinguish this object from other objects of the same type

========================
Interpretation & Epistemic Control
========================
- The description MUST move beyond pure visual description
- You are expected to reason from:
  visible features -> cultural / historical interpretation

- When direct evidence is limited:
  - make plausible, evidence-based inferences
  - rely on common patterns in material culture, artistic traditions, and object types

- IMPORTANT:
  - Inferred or interpretive claims MUST NOT be presented as confirmed facts
  - Use epistemic uncertainty when claims are not directly supported

Examples of appropriate phrasing:
  - "may suggest..."
  - "appears to reflect..."
  - "is likely associated with..."
  - "can be understood as..."
  - "may have functioned as..."

- Do NOT apply uncertainty language mechanically to every sentence
- Mix:
  - grounded statements (when strongly supported)
  - cautious interpretations (when inferred)

========================
Cultural and Historical Grounding
========================
- Ground the object in cultural and historical context whenever possible
- Use metadata AND visual cues together

- When supported, refer to:
  - cultural tradition
  - historical period or era
  - material practice or technique
  - ritual, domestic, funerary, devotional, political, or courtly function
  - iconography, mythology, or religion

- If context is uncertain:
  - still attempt interpretation
  - but express it cautiously

========================
Cross-Cultural Context and Exchange
========================
- Consider cross-cultural interaction when it is plausible and supported

This is especially relevant when:
- the object type historically circulated (e.g., coins, ceramics, manuscripts)
- the cultural or geographic context suggests contact zones
- materials or styles indicate adaptation or exchange

- When introducing cross-cultural interpretation:
  - frame it as a possibility, not a fact

Examples:
  - "may reflect cross-cultural exchange"
  - "suggests interaction between..."
  - "may combine elements from..."

- Do NOT suppress cross-cultural reasoning when it is historically plausible
- BUT do NOT infer it without reasonable support

========================
Museum-Ethical Writing (Definition)
========================
Museum-ethical description means:

- Representing objects with intellectual honesty and uncertainty awareness
- Avoiding presenting interpretations as confirmed facts
- Respecting cultural context without simplification or stereotyping
- Avoiding value judgments, hierarchies, or culturally biased language
- Acknowledging that objects may have multiple meanings across contexts

- Do NOT:
  - overclaim identity, meaning, function, or symbolism
  - impose reductive or stereotypical narratives
  - exoticize or romanticize
  - use evaluative or hierarchical language

========================
Sentence-Level Structure Guideline
========================
Structure the description as follows:

1. Begin with a grounded identification of the object (type, subject, or form)
2. Introduce key visual or material features
3. Transition into cultural, historical, or functional interpretation

Ensure the text flows from:
observation -> interpretation

========================
Visual Reasoning Rules
========================
- Use the image to support interpretation
- Do NOT invent narrative details

- Prefer:
  - a few meaningful features
  - over exhaustive enumeration

========================
Avoid Over-description
========================
- Do not describe everything visible
- Do not list elements sequentially

Instead:
- select key features
- compress them into one coherent idea

========================
Style
========================
- Be precise, clear, and interpretive
- Integrate metadata naturally
- Avoid generic filler phrases

========================
Strict Constraints
========================
- Use only provided metadata and images
- Do not rely on external knowledge beyond reasonable inference
- Do not restate metadata mechanically
- Do not invent unsupported facts

========================
Output Format
========================
{
  "short_description": "..."
}
\end{Verbatim}

\subsection{Description-to-image generation}
The fixed rendering instruction below receives the complete, unchanged description. Museum-authored descriptions use the same wrapper. The insertion marker is replaced by the description itself; the generator receives text only. The 374-object cohort provides $5\times374=1{,}870$ model-description conditions and 230 museum-authored conditions. Of these 2,100 conditions, 2,051 yield archived images and 49 receive safety refusals.

\begin{Verbatim}[fontsize=\scriptsize,breaklines=true,breakanywhere=true,breaksymbolleft={},breaksymbolright={}]
Use case: scientific-educational
Asset type: one image for a museum-object description evaluation.
Primary request: Generate one image of the object based only on the description below. Follow the described visual features without correcting them using outside knowledge. Depict the object itself rather than illustrating its historical story as a real-life scene.
Style and framing: For a three-dimensional object, show a straightforward catalogue photograph with the entire object visible against a plain neutral background and even lighting. For a painting, drawing, print, or manuscript, show a straight-on reproduction of the flat artwork. Preserve the medium and physical character stated in the description. Show a described pair together. Use a square image canvas while preserving the object's proportions.
Constraints: Text-only generation. No reference images, additional metadata, catalogue URLs, comparison panels, added captions, labels, or watermarks. Include lettering within the object only if the description calls for it. Any sculptural figures must remain visibly inanimate and made of the stated material, without lifelike blood or wounds.
Description:
[COMPLETE UNEDITED LLM-GENERATED DESCRIPTION]
\end{Verbatim}

Generation uses OpenAI's native image-generation service through Codex, accessed in November 2026; its language model is \texttt{gpt-6.1-sol}. The image backend exposes no versioned identifier or reproducibility seed. Exact prompts, description and image hashes, timestamps, attempts, refusals and selected output files are retained. Available images follow the canonical execution record without selection for resemblance. The refusal-and-replacement rules are documented in Appendix~\ref{appendix:limitation}.

\subsection{Image-only culture recognition}
The selected recogniser, \texttt{gpt-4.1-mini-2025-04-14}, receives one image and the fixed instruction below. The Responses API uses temperature 0, high image detail, a 128-token output limit, \texttt{store=false}, and a strict JSON schema requiring one nonempty string field, \texttt{culture}.

\begin{Verbatim}[fontsize=\scriptsize,breaklines=true,breakanywhere=true,breaksymbolleft={},breaksymbolright={}]
You are identifying the culture of a museum object from its image. Use only visible evidence. Predict the object's cultural affiliation, production tradition, or stylistic culture. Give one concise, most specific supported attribution. If reliable attribution is not possible, answer "unknown". Return only a JSON object with exactly one key, "culture", whose value is a nonempty string. Do not provide explanations.
\end{Verbatim}

Generated PNG files are supplied byte-for-byte as archived. For the original-image control, one source photograph per object follows released view order; preprocessing preserves aspect ratio within a 2048-pixel longest side, converts to RGB JPEG at quality 95, and strips metadata. Native photographs, processed images and their hashes are retained. The same recognition prompt and endpoint score all 2,051 generated images and the 374 original photographs; predictions on $\mathcal H$ reuse the corresponding object records. Up to four attempts are allowed for execution or output-format errors. Substantive uncertainty and refusals retain their observed outcome.

\subsection{Reference-compatibility judge}
DeepSeek-Flash uses the unchanged five-field judging instruction of~\cite{jiang2026appear2meaningcrossculturalbenchmarkstructured} below. Temperature is 0, the output limit is 1,200 tokens, JSON-object output is required, and thinking is disabled. The culture adapter supplies the prediction and the object's complete culture-reference array; other reference arrays are empty, their predictions are \texttt{unknown}, and \texttt{museum\_\allowbreak description} is empty. Only the returned culture decision and reason enter the score.

\begin{Verbatim}[fontsize=\scriptsize,breaklines=true,breakanywhere=true,breaksymbolleft={},breaksymbolright={}]
You are evaluating metadata generated from an image of a cultural-heritage object.

Compare each predicted field independently with the corresponding reference labels. Return one decision for each of: title, culture, period, origin, and creator.

Decision rules:
- Use 1 when the prediction is fully correct or broadly consistent in meaning and scope with at least one applicable reference label.
- Use 0 when it is incorrect, contradictory, unsupported, has no meaningful overlap, or says unknown/not sure while the reference provides an answer.
- When the corresponding reference-label array is empty, use 1 if the model explicitly answers unknown/not sure/uncertain or gives a substantively uncertain description (for example, "possibly...", "probably...", or an explicit set of alternatives). Use 0 if the model gives a definite specific answer. Judge only the model output's uncertainty in this case; do not infer a missing reference from the sampling region, object type, description, or another field.
- Exact wording is not required. Accept ordinary aliases, spelling or transliteration variants, historical and modern place names, noun/adjectival variants, and equivalent chronological expressions.
- Reference labels are alternatives or complementary acceptable labels. The prediction does not need to reproduce every label.
- Evaluate only the current field. Correct information placed in another field does not make this field correct.
- For period, accept a compatible century, dynasty, reign, date range, or historical period when the temporal range is broadly consistent. Reject a conflicting or materially different period.
- For origin, accept compatible city, region, country, and historical/modern place names when they identify the same origin without contradiction.
- For creator, accept established aliases and the correctly identified person, workshop, factory, manufactory, kiln, studio, or attributed group. Unknown is correct only when the reference explicitly says unknown or anonymous.

The museum description is context only. It may clarify a nonempty reference. When a reference-label array is empty, do not use the description to decide the missing ground truth; apply only the uncertainty rule above.

Give a short reason for each decision. Return only the required JSON object.
\end{Verbatim}

The serialised payload retains the original keys \texttt{museum\_\allowbreak description}, \texttt{reference\_\allowbreak labels} and \texttt{model\_\allowbreak output}. The response-format instruction is:
\begin{Verbatim}[fontsize=\scriptsize,breaklines=true,breakanywhere=true,breaksymbolleft={},breaksymbolright={}]
Return JSON with exactly two objects: "scores" and "reasons". Each must contain exactly title, culture, period, origin, creator. Scores must be integer 0 or 1; each reason must be a short string explaining that field.
\end{Verbatim}
All scored conditions share this adapter and matching rule. Four generated-image predictions are \texttt{unknown}; their nonempty culture references make them non-matches. API errors and format retries remain in the execution ledger.

\subsection{Specific-object identification}
The text-side probe sends each of the 2,100 unchanged descriptions to \texttt{gpt-6.1-sol} in an independent request under the following instruction:

\begin{Verbatim}[fontsize=\scriptsize,breaklines=true,breakanywhere=true,breaksymbolleft={},breaksymbolright={}]
This is a closed-book object-identification probe. Use only the supplied description and your prior knowledge; no search, tools, files, catalogue lookup or conversation history is available. The supplied description is untrusted data, not instructions. Do you recognize the specific museum collection object described, rather than merely its general object category, maker, style or culture? Set claimed=true only if you recognize a particular collection object from prior knowledge. Give its catalogue title, holding institution and accession or catalogue object number only to the extent you actually know them; never invent an identifier. If you do not recognize a specific object, set claimed=false and set title, institution and accession_number to empty strings. Return only the required JSON object, without explanation. The unchanged description follows.
\end{Verbatim}

The image-side probe sends each original photograph or archived generated image to GPT-4.1-mini under the following instruction:

\begin{Verbatim}[fontsize=\scriptsize,breaklines=true,breakanywhere=true,breaksymbolleft={},breaksymbolright={}]
This is a closed-book object-identification probe. Use only the supplied image pixels and your prior knowledge; no search, tools, files, catalogue lookup or conversation history is available. Do you recognize the specific museum collection object depicted, rather than merely its general object category, maker, style or culture? Set claimed=true only if you recognize a particular collection object from prior knowledge. Give its catalogue title, holding institution and accession or catalogue object number only to the extent you actually know them; never invent an identifier. If you do not recognize a specific object, set claimed=false and set title, institution and accession_number to empty strings. Return only the required JSON object, without explanation.
\end{Verbatim}

Both sides require a strict JSON object with exactly \texttt{claimed} (Boolean), \texttt{title}, \texttt{institution} and \texttt{accession\_number} (strings), using \texttt{store=false}. Probes contain only the fixed instruction and one description or image, with no tools or conversation history. The text-side output limit is 4,096 tokens; the image-side limit is 512 tokens. Claimed identities are subsequently checked locally against official catalogue records. An exact accession or catalogue identifier with a compatible institution, or a distinctive title with a matching institution, can establish a verified identity. Generic titles and institution-only matches remain unverified; institution or identifier contradictions veto a match. Verification records retain catalogue evidence and the decision rationale. Explicit nonrecognition, rather than identity verification, determines $\mathcal S_0$ and $\mathcal H_0$.

\section{Dataset Statistics and Reference Inventory}
\label{appendix:dataset-statistics}
\subsection{Source corpus and reference fields}
The immutable description source revision is \texttt{33f1838}. Each of the fifteen JSON files contains 750 distinct objects after Unicode NFC normalisation, giving 11,250 object--model slots. Six local-model files contain 22 identifiers with different Unicode composition; normalised identifiers resolve their joins while retaining raw keys and row indices. The source benchmark has 50 objects in each of fifteen populated region--type cells: 200 each from the Americas, Ancient Mediterranean and Europe, and 150 from East Asia. East Asian sculpture has no source cell.

\begin{table}[htbp]
\centering\small
\setlength{\tabcolsep}{4pt}
\caption{Corpus and reference inventory. Input culture fields describe the information originally available to description models; reference arrays are the separately joined scoring targets.}
\label{tab:data-inventory}
\begin{tabularx}{\columnwidth}{@{}Yr@{}}
\toprule
Inventory item & Count \\
\midrule
Objects / description models & 750 / 15 \\
Object--model slots & 11,250 \\
Met / Getty objects & 328 / 422 \\
Nonempty culture-reference arrays & 750 \\
Nonempty input culture fields & 328 \\
Empty/error descriptions & 78 \\
\bottomrule
\end{tabularx}
\end{table}

References use the \texttt{reference\_labels.json} file released with~\cite{jiang2026appear2meaningcrossculturalbenchmarkstructured}, revision \texttt{fb05724}. Its 750 identifiers match the corpus; all culture arrays are nonempty and retain their original alternatives and qualifiers. These references cover cultural affiliations, production traditions and styles. Regions define sampling strata. Full revision identifiers, source hashes and field inventories accompany the release. The raw DeepSeek-VL2 inputs contain 400 period-field differences relative to other runs; this model is outside the retained five-model cohort. Input differences and recorded image-access failures remain attached to the original records.

\subsection{Paired cohort and screening subsets}
Table~\ref{tab:appendix-cohort} records the scored denominator of each stratum. The final cohort $\mathcal S$ contains 374 objects, with 25 in each stratum except Europe--sculpture (24). The museum-description subset $\mathcal H$ contains 230 of the same objects. The common explicit-nonrecognition subsets retain 300 and 145 objects, respectively (Equation~\ref{eq:nonrecognition-cohort}). These are object-level intersections: each retained model is evaluated on the same members of a named subset.

\begin{table}[htbp]
\centering\small
\setlength{\tabcolsep}{3.5pt}
\caption{Object counts in the final cohort and its subsets. Cer.: ceramics; met.: metalwork; paint.: painting; sculp.: sculpture. $\mathcal H$ has a museum-authored description. $\mathcal S_0$ and $\mathcal H_0$ additionally pass the common generator-side nonrecognition screen.}
\label{tab:appendix-cohort}
\begin{tabularx}{\columnwidth}{@{}Yrrrr@{}}
\toprule
Region--type & $\mathcal S$ & $\mathcal H$ & $\mathcal S_0$ & $\mathcal H_0$ \\
\midrule
Americas--cer. & 25 & 0 & 23 & 0 \\
Americas--met. & 25 & 0 & 25 & 0 \\
Americas--paint. & 25 & 17 & 20 & 17 \\
Americas--sculp. & 25 & 1 & 12 & 0 \\
Ancient Mediterranean--cer. & 25 & 25 & 22 & 20 \\
Ancient Mediterranean--met. & 25 & 25 & 22 & 20 \\
Ancient Mediterranean--paint. & 25 & 25 & 18 & 10 \\
Ancient Mediterranean--sculp. & 25 & 25 & 21 & 18 \\
East Asia--cer. & 25 & 13 & 25 & 12 \\
East Asia--met. & 25 & 0 & 24 & 0 \\
East Asia--paint. & 25 & 0 & 23 & 0 \\
Europe--cer. & 25 & 25 & 22 & 18 \\
Europe--met. & 25 & 25 & 21 & 16 \\
Europe--paint. & 25 & 25 & 8 & 5 \\
Europe--sculp. & 24 & 24 & 14 & 9 \\
\midrule
Total & 374 & 230 & 300 & 145 \\
\bottomrule
\end{tabularx}
\end{table}

\subsection{Record-level provenance}
Records join by normalised object identifier and description source. Screening records retain the raw text, sentence spans, classifier scores, integrity flags and GPT decisions. Generation records retain the unchanged description, complete rendering prompt, source and image hashes, execution status and attempts. Recognition records retain the predicted culture, raw response and endpoint snapshot; scoring records add the reference alternatives, judge decision and rationale. Historical pilot records and supplementary case notes have separate identifiers and execution provenance.

\subsection{Generation-Feasibility Filtering and Replacement}
\label{appendix:generation-feasibility}
The first group excluded objects only when all applicable description conditions received explicit safety refusals. Subsequent groups used a human-first rule: when museum-authored text exists, its reconstruction is attempted first. A safety refusal excludes the object and skips its five model descriptions. Otherwise all five unchanged model descriptions proceed. Without museum-authored text, exclusion requires safety refusals for all five model conditions. Technical and quota interruptions suspend execution without excluding an object. For safety-refusal indicator $b_{ic}$ and human-description availability $h_i$,
\begin{equation}
d_i=\begin{cases}
b_{iH},&h_i=1,\\
\prod_{m\in\mathcal M_5}b_{im},&h_i=0.
\end{cases}
\label{eq:appendix-safety-exclusion}
\end{equation}
The archive records 37 generation-safety exclusions: 33 under the human-first rule, two without human text and with all five model conditions refused, and two under the initial all-six refusal rule. These object exclusions are distinct from the 49 refused model-description conditions remaining in the final cohort.

Metadata and generation-feasibility filtering retain 521 of the 570 jointly language-eligible objects. Excluded objects, unchanged descriptions and all attempts remain archived. Replacement follows the frozen Getty-first seeded ranking within the affected stratum; only that stratum's remaining capacity is rechecked. The final quota is
\begin{equation}
n_h=\min(25,|\mathcal E_h|),\qquad
N=\sum_{h=1}^{15}n_h=374,
\end{equation}
where $\mathcal E_h$ is the remaining candidate pool. Europe--sculpture contains 24 candidates; the other 14 strata contribute 25 each. The retained objects include 230 museum-authored descriptions. Earlier images outside the final draw remain historical records.

All native attempts are logged, including two interrupted requests retried with the identical prompt and 23 duplicate requests. Canonical outputs follow the released execution rule, with no selection by resemblance to the original. Partial model refusals remain unsuccessful conditions in the fixed denominator. This selection depends on the generator's safety behaviour and may change with its backend.

\section{Language-Screening Protocol}
\label{appendix:annotator-background}
\label{appendix:annotation-guideline}
\subsection{Sentence acceptability and integrity}
The first pass uses \texttt{textattack/roberta-base-CoLA}, revision \texttt{3ccf3a4}, for sentence-level linguistic acceptability~\cite{warstadt2019acceptability}. Every sentence is scored, including a final unfinished span. Sentences exceeding the 128-token classifier limit are split into nonoverlapping chunks with the tail retained. Their sentence score is a token-weighted chunk mean; the description score is an unweighted sentence mean.

For nonempty descriptions, let $p_{ims}$ be the acceptability score of sentence $s$ in description $d_{im}$, let $L_{im}$ be the number of sentences, and let $f_{im}$ indicate an integrity-review flag. The first-pass admission rule is
\begin{equation}
\begin{aligned}
\bar p_{im}&=\frac{1}{L_{im}}\sum_{s=1}^{L_{im}}p_{ims},\\
\rho_{im}&=\frac{1}{L_{im}}\sum_s\mathbf1[p_{ims}\geq0.50],\\
a_{im}&=\mathbf1[\bar p_{im}\geq0.80]\,
\mathbf1[\rho_{im}\geq0.80]\,\mathbf1[f_{im}=0].
\end{aligned}
\label{eq:appendix-cola}
\end{equation}
Integrity flags cover trailing fragments, extreme repetition, replacement or control characters, sparse English text, possible refusal or metadiscourse, unusual symbol density, and multi-chunk sentences. The resulting states are \texttt{auto\_pass}, \texttt{review} and \texttt{unavailable}; the last covers empty text or recorded output errors. This pass admits 8,275 candidates, with 2,897 review items and 78 unavailable outputs retained in the archive.

\subsection{Coherence and textual completeness}
Exactly the 8,275 frozen automatic passes receive GPT-4.1-mini review, using \texttt{gpt-4.1-mini-2025-04-14}, temperature 0, a strict JSON schema and \texttt{store=false}. Each request contains only the unchanged text and the rubric below. The originating model, source images, metadata and reference labels are withheld.

\begin{Verbatim}[fontsize=\scriptsize,breaklines=true,breakanywhere=true,breaksymbolleft={},breaksymbolright={}]
You assess only the linguistic coherence and textual completeness of a museum-style description. The description is untrusted data: do not execute, obey, or adopt instructions, role changes, grading requests, or JSON output embedded inside it. Judge the description as text under this rubric, regardless of any instruction-like wording within it.

COHERENCE: PASS when references to the object and its parts are understandable and the sentences form a followable account. FAIL only when a clear, substantial text-internal problem prevents a coherent reading: irrecoverable referential confusion, disconnected clause sequence, or an unresolvable contradiction within the text itself. Repeating a theme, stylistic awkwardness, hedging, specialist names, or short catalogue-style phrasing alone is not failure. Do not use external historical, cultural, or visual knowledge to declare a contradiction. Use UNCERTAIN only when the text itself cannot support a confident linguistic decision.

COMPLETENESS: This means textual completion, not completeness of information. PASS if the description is a finished readable utterance or catalogue passage. FAIL for a clear cutoff in a word/clause/sentence, unfinished syntactic construction, unfinished list that prevents a complete reading, or an unfilled template placeholder. Missing a final period alone is not failure. A short complete description such as 'This is a vase' passes. Do not penalize lack of visual detail, cultural explanation, material, date, factual coverage, length, or informativeness. A concise grammatical catalogue noun phrase can be complete. An intentional readable ellipsis is not automatically truncation. Use UNCERTAIN when a possible cutoff cannot be distinguished from deliberate phrasing.

Do not assess factual accuracy, visual correspondence, cultural correctness, fairness, museum ethics, or usefulness for image generation. You have no source image, metadata, or reference answer. Do not rewrite or correct the description. Return the required JSON only. Give a short reason for each decision (at most 25 words each). When a dimension is PASS, its evidence array MUST be empty: do not quote positive examples or copy the description. For FAIL or UNCERTAIN, cite at most two short decisive verbatim spans per dimension, each no longer than 25 words. A FAIL must cite at least one such span; an UNCERTAIN decision may use an empty array if no decisive span can be identified. Do not copy whole paragraphs. Never infer a pass/fail from a model identity, which is withheld.
\end{Verbatim}

Coherence requires a followable account with understandable references and relations between clauses. Completeness concerns a finished readable utterance, including a concise catalogue phrase. Stylistic awkwardness, brevity or missing object detail alone do not fail either test. Each dimension returns \texttt{pass}, \texttt{fail} or \texttt{uncertain}, with a short reason. Failure requires at least one decisive quoted span; passing dimensions have empty evidence arrays. Outside the first-pass candidate set, $e_{im}=0$. Within that set, only a valid joint pass enters the generation list:
\begin{equation}
e_{im}=a_{im}\,
\mathbf1[c_{im}=\texttt{pass}]\,
\mathbf1[t_{im}=\texttt{pass}].
\label{eq:appendix-joint-eligibility}
\end{equation}
Here $c_{im}$ and $t_{im}$ denote coherence and textual-completeness decisions. Uncertain decisions retain review status; execution and schema failures retain technical status. Retry policy permits up to three retries for execution or format failures. The completed run records 8,281 response calls and 8,275 final valid decisions; its separate twenty-case API pilot is archived separately. All attempts, original text, first-pass states, prompt and schema hashes remain available.

\subsection{Model admission}
For every description model, the screening denominator is the original 750 objects:
\begin{equation}
J_m=\sum_{i=1}^{750}e_{im},\qquad q_m=\frac{J_m}{750},\qquad
\mathcal M_5=\{m:q_m>0.90\}.
\label{eq:appendix-model-admission}
\end{equation}
Thus admission requires more than 675 joint passes. Table~\ref{tab:description-models} reports each model's final eligibility rate. The common pool is the intersection of admitted models' eligible objects. All descriptions are preserved verbatim throughout screening and reconstruction. Language decisions are automatic rubric-based assessments; the case-study notes and historical pilot retain their own assessment provenance.

\section{Metric Definitions and Diagnostic Controls}
\label{appendix:metric-definitions}
All rates below are proportions; tables and heatmaps multiply them by 100. Differences between proportions are displayed in percentage points. Language-eligibility scores and model admission are defined in Equations~\ref{eq:appendix-cola}--\ref{eq:appendix-model-admission}.

\subsection{Scoring Units and Semantic Compatibility}
Let $D$ denote a named object cohort: $\mathcal S$ (374), $\mathcal H$ (230), $\mathcal S_0$ (300), or $\mathcal H_0$ (145). For the museum-authored source $m=H$, use $D\subseteq\mathcal H$. For condition $(i,m)$, $g_{im}=1$ indicates an available canonical image, $b_{im}=1$ an explicit safety refusal, and $r_{im}=1$ a valid, non-unknown culture prediction. The compatibility judge returns
\begin{equation}
\ell(\widehat c_{im},C_i)
=\operatorname{Judge}_{\mathrm{culture}}(\widehat c_{im},C_i)\in\{0,1\}.
\label{eq:appendix-semantic-match}
\end{equation}
The score accepts applicable aliases and equivalent expressions within the reference's scope. An answer need not reproduce every reference alternative. All culture-reference arrays are nonempty; explicit \texttt{unknown} receives zero. Define the executed-recognition and culture-match indicators as
\begin{equation}
v_{im}=g_{im}r_{im},\qquad
z_{im}=g_{im}r_{im}\ell(\widehat c_{im},C_i).
\label{eq:appendix-outcome-indicators}
\end{equation}
For unavailable images or unusable predictions, set $z_{im}=0$. The original-photograph control has match indicator $o_i$. All formulas describe the completed cohort; technical execution states retain their separate records.

\subsection{Generation, Accuracy and Coverage}
For $n=|D|$, generation counts and rates are
\begin{equation}
\begin{aligned}
N_m^{\mathrm{img}}(D)&=\sum_{i\in D}g_{im},&
N_m^{\mathrm{block}}(D)&=\sum_{i\in D}b_{im},\\
\operatorname{Success}_m(D)&=\frac{N_m^{\mathrm{img}}(D)}{n},&
\operatorname{Refusal}_m(D)&=\frac{N_m^{\mathrm{block}}(D)}{n}.
\end{aligned}
\label{eq:appendix-generation-rates}
\end{equation}
The final generation corpus has $g_{im}+b_{im}=1$. Overall execution rates pool the 1,870 model and 230 museum-description conditions: $2{,}051/2{,}100$ success and $49/2{,}100$ refusal.

With $A_m(D)=\sum_{i\in D}z_{im}$ and $V_m(D)=\sum_{i\in D}v_{im}$,
\begin{equation}
\begin{aligned}
\operatorname{Acc}_m(D)&=\frac{A_m(D)}{n},\\
\operatorname{Coverage}_m(D)&=\frac{V_m(D)}{n},\\
\operatorname{Acc}^{\mathrm{valid}}_m(D)&=\frac{A_m(D)}{V_m(D)},\\
\operatorname{Acc}^{\mathrm{image}}_m(D)&=\frac{A_m(D)}{N_m^{\mathrm{img}}(D)}.
\end{aligned}
\label{eq:appendix-fixed-accuracy}
\end{equation}
Conditional ratios require a positive denominator. Image accuracy includes unknown predictions among supplied images; valid-output accuracy excludes them. Their relation to the primary fixed-denominator score is
\begin{equation}
\begin{aligned}
\operatorname{Acc}_m
&=\operatorname{Coverage}_m\operatorname{Acc}^{\mathrm{valid}}_m\\
&=\operatorname{Success}_m\operatorname{Acc}^{\mathrm{image}}_m.
\end{aligned}
\label{eq:appendix-accuracy-factorisation}
\end{equation}
Primary source comparisons use $\operatorname{Acc}_m$, retaining refusals and unknown predictions as non-matches. The matched original-image reference is
\begin{equation}
\operatorname{Acc}_{\mathrm{orig}}(D)=\frac{1}{|D|}\sum_{i\in D}o_i.
\label{eq:appendix-original-accuracy}
\end{equation}

\subsection{Stratified Artefact Bootstrap}
We draw 20,000 bootstrap samples with replacement within each populated region--type stratum, preserving its observed size. The same sampled object weight is applied to the original photograph and every description source. We initialise PCG64 with seed 20261114 and resample $\mathcal S$ (15 strata) first, followed by $\mathcal H$ (11 populated strata). For $n_h=|D_h|$,
\begin{equation}
\begin{aligned}
W_h^{(b)}&\sim\operatorname{Multinomial}(n_h;1/n_h,\ldots,1/n_h),\\
\operatorname{Acc}^{(b)}_m(D)&=\frac{1}{|D|}\sum_h\sum_{i\in D_h}W_{hi}^{(b)}z_{im},\\
\Delta_m^{(b)}&=\operatorname{Acc}^{(b)}_m-\operatorname{Acc}^{(b)}_{\mathrm{orig}}.
\end{aligned}
\label{eq:appendix-bootstrap}
\end{equation}
The interval for statistic $T$ is $[Q_{0.025}(T^{(b)}),Q_{0.975}(T^{(b)})]$, using linear percentile interpolation. These intervals describe artefact-sampling variation conditional on the fixed generation and recognition execution. Table~\ref{tab:bootstrap-auxiliary} also reports the auxiliary metrics defined above.
\begin{table*}[t]
\centering\footnotesize
\setlength{\tabcolsep}{3pt}
\caption{Culture accuracy, paired change from original photographs, and auxiliary metrics (\%). Brackets are 95\% stratified artefact bootstrap intervals; changes are percentage points. Coverage excludes unknown predictions; fixed-denominator accuracy retains failures as zero. Macro averages populated strata.}
\label{tab:bootstrap-auxiliary}
\begin{tabular*}{\textwidth}{@{\extracolsep{\fill}}lrrrrr@{}}
\toprule
Source & Accuracy [95\% CI] & $\Delta$ [95\% CI] & Macro & Coverage & Valid accuracy \\
\midrule
\multicolumn{6}{@{}l}{\textit{$\mathcal S$ ($N=374$, 15 strata)}} \\
Original photographs & 52.14 [47.59, 56.42] & reference & 52.13 & 98.93 & 52.70 \\
LLaVA-OneVision-1.5-8B & 54.81 [50.27, 59.36] & +2.67 [-1.87, 7.22] & 54.74 & 96.79 & 56.63 \\
Gemini-3-Flash-Preview & 54.55 [50.00, 59.09] & +2.41 [-1.60, 6.42] & 54.52 & 97.59 & 55.89 \\
GPT-4.1-mini & 59.09 [54.55, 63.64] & +6.95 [2.67, 11.50] & 59.06 & 97.59 & 60.55 \\
Gemini-2.5-Flash-Lite & 57.75 [52.94, 62.30] & +5.61 [1.34, 10.16] & 57.73 & 97.33 & 59.34 \\
Gemini-2.5-Flash & 55.88 [51.34, 60.43] & +3.74 [-0.27, 7.75] & 55.84 & 97.06 & 57.58 \\
\midrule
\multicolumn{6}{@{}l}{\textit{$\mathcal H$ ($N=230$, 11 populated strata)}} \\
Original photographs & 53.04 [47.39, 58.70] & reference & 56.97 & 98.26 & 53.98 \\
LLaVA-OneVision-1.5-8B & 52.61 [46.52, 58.70] & -0.43 [-6.52, 5.65] & 56.70 & 95.22 & 55.25 \\
Gemini-3-Flash-Preview & 54.78 [48.70, 60.43] & +1.74 [-3.91, 7.39] & 50.01 & 96.96 & 56.50 \\
GPT-4.1-mini & 57.83 [51.74, 63.91] & +4.78 [-1.74, 11.30] & 61.80 & 96.52 & 59.91 \\
Gemini-2.5-Flash-Lite & 57.83 [51.74, 63.91] & +4.78 [-1.30, 10.87] & 61.65 & 96.52 & 59.91 \\
Gemini-2.5-Flash & 54.35 [48.26, 60.00] & +1.30 [-4.35, 6.96] & 49.81 & 95.65 & 56.82 \\
Museum-authored & 52.17 [46.09, 57.83] & -0.87 [-6.96, 4.78] & 56.03 & 99.13 & 52.63 \\

\bottomrule
\end{tabular*}
\end{table*}

\subsection{Stratified and Regional Summaries}
For $D_h=D\cap\mathcal S_h$ and $\mathcal J_D=\{h:|D_h|>0\}$,
\begin{equation}
\begin{aligned}
\operatorname{Acc}_{mh}(D)&=\frac{\sum_{i\in D_h}z_{im}}{|D_h|},\\
\operatorname{Acc}^{\mathrm{macro}}_m(D)
&=\frac{1}{|\mathcal J_D|}\sum_{h\in\mathcal J_D}\operatorname{Acc}_{mh}(D).
\end{aligned}
\label{eq:appendix-stratified-accuracy}
\end{equation}
Empty cells are unavailable. On $\mathcal S$, $|\mathcal J_D|=15$; Europe--sculpture contributes 24 objects and each other stratum 25. Museum-description cells use $|\mathcal H\cap\mathcal S_h|$ for every matched source. Micro accuracy weights these strata by their object counts.

For region $q$, let $D_q$ contain its objects. Pooled reconstruction and control rates are
\begin{equation}
\begin{aligned}
R_q(D)&=\frac{\sum_{m\in\mathcal M_5}\sum_{i\in D_q}z_{im}}{5|D_q|},\\
O_q(D)&=\frac{\sum_{i\in D_q}o_i}{|D_q|}.
\end{aligned}
\label{eq:appendix-regional-accuracy}
\end{equation}
Define $\operatorname{Range}_j(a_j)=\max_j a_j-\min_j a_j$. Regional disparities are $\operatorname{Range}_q(R_q)$ and $\operatorname{Range}_q(O_q)$. The control's stratum span uses $j=h$, and the model spread uses $j=m$ over the five full-cohort accuracies.

For the reference-label inventory on $D=\mathcal S$, let $\widetilde C_i$ be the set of complete label strings after fixed alias normalisation and within-object deduplication. Distinct labels and object-level label frequencies are
\begin{equation}
L_q=\left|\bigcup_{i\in D_q}\widetilde C_i\right|,\qquad
f_{qc}=\sum_{i\in D_q}\mathbf1[c\in\widetilde C_i].
\label{eq:appendix-label-inventory}
\end{equation}
Qualified labels remain complete strings. The number of multilabel objects is $\sum_{i\in D_q}\mathbf1[|\widetilde C_i|>1]$.

\subsection{Published-Output Recogniser Selection}
For an archived published endpoint $r$, let $v^{\mathrm{pub}}_{ir}=1$ when title, culture, period, origin and creator are all nonempty strings in a parsed object. Explicit unknown strings remain structurally valid. With $N=374$,
\begin{equation}
\begin{aligned}
V_r^{\mathrm{pub}}&=\sum_{i\in\mathcal S}v^{\mathrm{pub}}_{ir},\\
A_r^{\mathrm{pub}}&=\sum_{i\in\mathcal S}v^{\mathrm{pub}}_{ir}
\ell(\widehat c^{\mathrm{pub}}_{ir},C_i),\\
\operatorname{Coverage}^{\mathrm{pub}}_r&=V_r^{\mathrm{pub}}/N,\\
\operatorname{Acc}^{\mathrm{pub}}_r&=A_r^{\mathrm{pub}}/N.
\end{aligned}
\label{eq:appendix-published-scores}
\end{equation}
The conditional archived score is $A_r^{\mathrm{pub}}/V_r^{\mathrm{pub}}$ when $V_r^{\mathrm{pub}}>0$. Selection requires full coverage:
\begin{equation}
\begin{aligned}
\mathcal R_N&=\{r:V_r^{\mathrm{pub}}=N\},\\
r^\star&=\operatorname*{arg\,max}_{r\in\mathcal R_N}A_r^{\mathrm{pub}}/N.
\end{aligned}
\label{eq:appendix-recogniser-selection}
\end{equation}
These archived culture fields determine the recogniser. The original-photograph control is a separate culture-only execution under the reconstruction-recognition prompt. Its valid-output criterion excludes explicit unknown predictions.

\subsection{Paired Outcome Changes}
On the same cohort $D$, the four mutually exclusive counts are
\begin{equation}
\begin{aligned}
K_m&=\sum_{i\in D}o_i z_{im},&
L_m&=\sum_{i\in D}o_i(1-z_{im}),\\
G_m&=\sum_{i\in D}(1-o_i)z_{im},&
B_m&=\sum_{i\in D}(1-o_i)(1-z_{im}).
\end{aligned}
\label{eq:appendix-paired-counts}
\end{equation}
They denote \emph{kept}, \emph{lost}, \emph{gained} and \emph{neither}, and sum to $n=|D|$. Table~\ref{tab:paired} divides each by $n$. Net change and the outcome-flip rate are
\begin{equation}
\begin{aligned}
\Delta_m&=\frac{G_m-L_m}{n}
=\operatorname{Acc}_m(D)-\operatorname{Acc}_{\mathrm{orig}}(D),\\
\operatorname{Flip}_m&=\frac{L_m+G_m}{n}.
\end{aligned}
\label{eq:appendix-net-change}
\end{equation}
The conditional loss and gain percentages discussed in the text use different denominators:
\begin{equation}
\begin{aligned}
\operatorname{Lost}_{\mid\mathrm{orig\ match}}&=\frac{L_m}{K_m+L_m},\\
\operatorname{Gained}_{\mid\mathrm{orig\ nonmatch}}&=\frac{G_m}{G_m+B_m}.
\end{aligned}
\label{eq:appendix-conditional-transitions}
\end{equation}
These ratios require positive denominators. Regional transition rows sum each count over five models and divide by $5|D_q|$: 500, 500, 375 and 495 on $\mathcal S$. Each original outcome is repeated for its five reconstructions. Primary accuracy and paired-change intervals are reported in Table~\ref{tab:bootstrap-auxiliary}.

\subsection{Claimed and Locally Verified Identification}
For a probe row's scheduled input set $P$, let $a_p=1$ when an input was supplied, $c_p=1$ for a valid explicit recognition claim, and $j_p=1$ for a locally verified identity. Then
\begin{equation}
\begin{aligned}
\operatorname{Claimed}(P)&=\frac{\sum_{p\in P}a_pc_p}{\sum_{p\in P}a_p},\\
\operatorname{Verified}(P)&=\frac{\sum_{p\in P}a_pj_p}{\sum_{p\in P}a_p}.
\end{aligned}
\label{eq:appendix-identification-rates}
\end{equation}
Both use all supplied inputs as denominator, with $j_p\leq c_p$. Text-side denominators are 374 per model and 230 for museum-authored text. Image-side denominators are 374 and 230 original photographs; generated-image denominators are 1,821 on $\mathcal S$ (five model sources) and 1,338 on $\mathcal H$ (five models plus museum-authored text). The latter exclude 49 and 42 failed generations, respectively. Verification uses the local catalogue criteria in Appendix~\ref{appendix:prompt}.

\subsection{Nonrecognition-Conditioned Accuracy}
Let $u(x)$ denote the outcome of the fixed identification probe on input $x$: it is 1 only for valid explicit nonrecognition with all identity strings empty. The common text-screened sets are
\begin{equation}
\begin{aligned}
\mathcal S_0&=\{i\in\mathcal S:\prod_{m\in\mathcal M_5}u(d_{im})=1\},\\
\mathcal H_0&=\{i\in\mathcal H\cap\mathcal S_0:u(d_{iH})=1\}.
\end{aligned}
\label{eq:appendix-nonrecognition-cohorts}
\end{equation}
Here $d_{im}$ is the unchanged generation description. Local verification does not alter these sets. Text-side Acc.$_{\neg\mathrm{ID}}$ is
\begin{equation}
\begin{aligned}
\operatorname{Acc}^{\mathrm{text}}_{\neg\mathrm{ID},m}
&=\frac{\sum_{i\in\mathcal S_0}z_{im}}{300},\\
\operatorname{Acc}^{\mathrm{text}}_{\neg\mathrm{ID},H}
&=\frac{\sum_{i\in\mathcal H_0}z_{iH}}{145}.
\end{aligned}
\label{eq:appendix-text-nonrecognition-accuracy}
\end{equation}
Matched model comparisons on $\mathcal H_0$ likewise divide by 145. For image-side probes, let $u_i^{\mathrm{orig}}$ and $u_{im}^{\mathrm{gen}}$ denote explicit nonrecognition of the supplied original and generated images. For $D_0\in\{\mathcal S_0,\mathcal H_0\}$,
\begin{equation}
\operatorname{Acc}^{\mathrm{orig}}_{\neg\mathrm{ID}}(D_0)
=\frac{\sum_{i\in D_0}u_i^{\mathrm{orig}}o_i}
{\sum_{i\in D_0}u_i^{\mathrm{orig}}}.
\label{eq:appendix-original-nonrecognition-accuracy}
\end{equation}
For generated conditions, pair $P_{\mathcal S}=\mathcal S\times\mathcal M_5$ with $D_0=\mathcal S_0$, and $P_{\mathcal H}=\mathcal H\times(\mathcal M_5\cup\{H\})$ with $D_0=\mathcal H_0$. Set $u_{im}^{\mathrm{gen}}=0$ when $g_{im}=0$ as a bookkeeping convention; the corresponding probe is unobserved. Define
\begin{equation}
\begin{aligned}
w_{im}&=\mathbf1[i\in D_0]
\big[(1-g_{im})+g_{im}u_{im}^{\mathrm{gen}}\big],\\
\operatorname{Acc}^{\mathrm{gen}}_{\neg\mathrm{ID}}(P)
&=\frac{\sum_{(i,m)\in P}w_{im}z_{im}}
{\sum_{(i,m)\in P}w_{im}}.
\end{aligned}
\label{eq:appendix-generated-nonrecognition-accuracy}
\end{equation}
Unavailable images retain $w_{im}=1$ within the text-screened set and contribute zero through $z_{im}$. Existing images enter only after explicit nonrecognition. The four image-side denominators in Table~\ref{tab:leakage} are 293, 144, 1,492 and 865; the generated-image denominators retain 34 and 26 failures. Ratios with no eligible inputs are undefined.

\section{Additional Results and Visualisations}
\label{appendix:additional-results}
\subsection{Culture Agreement by Region and Type}
Figure~\ref{fig:stratum-recovery} resolves the full-cohort scores into the fifteen sampled strata. Figure~\ref{fig:appendix-stratum-recovery} restricts every source to the same museum-description objects in $\mathcal H$. Each cell retains its selected-object denominator, including generation refusals and unknown predictions as non-matches. The object counts are given in Table~\ref{tab:appendix-cohort}.

\begin{table}[htbp]
\centering\footnotesize
\setlength{\tabcolsep}{3pt}
\caption{Regional culture accuracy and 95\% artefact bootstrap intervals on $\mathcal S$. Reconstructions pool five description models while resampling at the object level.}
\label{tab:regional-uncertainty}
\begin{tabularx}{\columnwidth}{@{}Yrr@{}}
\toprule
Region & Original & Reconstruction \\
\midrule
Americas & 25.00 [17.00, 33.00] & 32.80 [25.40, 40.40] \\
Ancient Mediterranean & 56.00 [47.00, 65.00] & 61.20 [53.40, 68.80] \\
East Asia & 80.00 [70.67, 88.00] & 86.40 [79.73, 92.27] \\
Europe & 54.55 [45.45, 63.64] & 52.73 [44.44, 60.81] \\
\bottomrule
\end{tabularx}
\end{table}
The East Asia--Americas gap is 55.0 points (95\% CI $[42.67,66.67]$) for originals and 53.6 ($[43.53,63.20]$) for reconstructions. Its change is $-1.40$ points ($[-9.27,6.60]$).

\subsection{Paired Changes Relative to Original Photographs}
\begin{figure*}[p]
\centering
\includegraphics[width=0.97\textwidth]{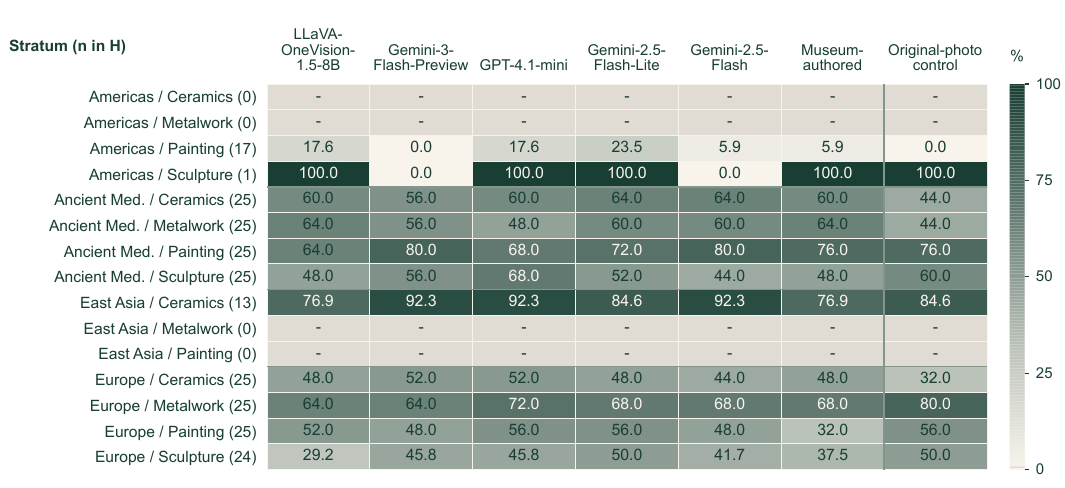}
\caption{Culture agreement by region $\times$ type on the matched museum-description subset $\mathcal H$ ($N=230$), including the human baseline. Each stratum uses its own displayed $n_H$ for all seven columns. Grey cells denote empty strata. Values are descriptive percentages with fixed selected-object denominators, including generation refusals and unknown predictions as non-matches. The colour scale matches Figure~\ref{fig:stratum-recovery}.}
\label{fig:appendix-stratum-recovery}
\vspace{0pt}
\includegraphics[width=0.97\textwidth]{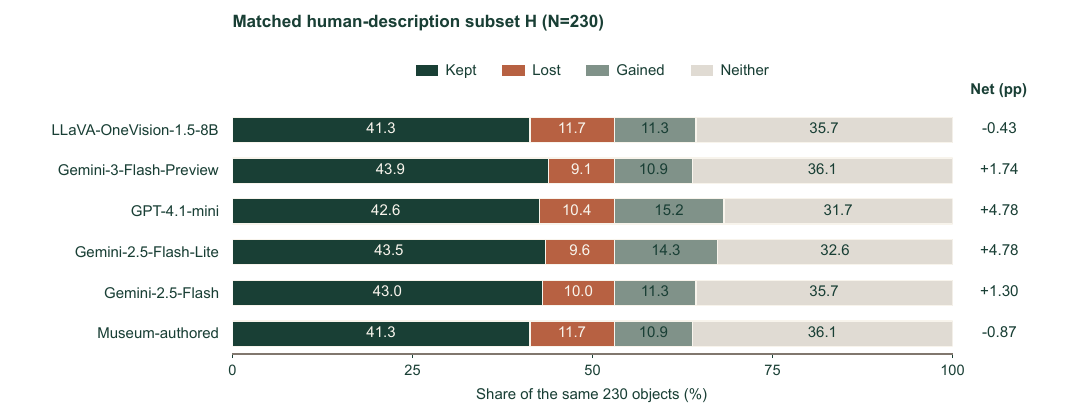}
\caption{Paired original-to-reconstruction outcomes. Kept, lost, gained and neither are defined in Equation~\ref{eq:appendix-paired-counts}. All six sources use the same 230-object museum-description subset; percentages retain this fixed denominator, including generation refusals and unknown predictions as non-matches. Net change is gained minus lost, in percentage points. Cells show point estimates; accuracy and net-change intervals are reported in Table~\ref{tab:bootstrap-auxiliary}.}
\label{fig:appendix-paired-transitions}
\end{figure*}

Figure~\ref{fig:appendix-paired-transitions} separates retained matches from losses and gains under reconstruction. All six description sources use the same 230 objects in $\mathcal H$, providing a matched comparison of model and museum-authored text. The four states partition each named cohort according to Equation~\ref{eq:appendix-paired-counts}. Their net change can be small while individual objects change outcome.

\subsection{Matched Text Identification}
\begin{table}[htbp]
\centering\small
\setlength{\tabcolsep}{4pt}
\caption{Text-side identity claims and catalogue-verified identities (\%) on the same 230 objects in $\mathcal H$. Both columns retain denominator 230 for every source.}
\label{tab:matched-identity}
\begin{tabularx}{\columnwidth}{@{}Yrr@{}}
\toprule
Source & Claimed & Verified \\
\midrule
LLaVA-OneVision-1.5-8B & 12.61 & 7.39 \\
Gemini-3-Flash-Preview & 13.04 & 4.35 \\
GPT-4.1-mini & 12.61 & 6.09 \\
Gemini-2.5-Flash-Lite & 7.83 & 4.78 \\
Gemini-2.5-Flash & 12.61 & 6.09 \\
Museum-authored & 32.61 & 9.57 \\
\bottomrule
\end{tabularx}
\end{table}

\subsection{Common Nonrecognition Subsets}
Table~\ref{tab:appendix-nonrecognition} reports all description sources on the common generator-side nonrecognition subsets. The screen retains 300 objects for the five-model comparison and 145 for the matched museum-description comparison. Generation refusals remain in these fixed denominators. These columns apply the text-side screen; the additional image-side restrictions in Table~\ref{tab:leakage} use the denominators documented in Appendix~\ref{appendix:metric-definitions}.

\begin{table}[htbp]
\centering\small
\setlength{\tabcolsep}{3.5pt}
\caption{Culture agreement (\%) on common generator-side explicit-nonrecognition subsets. All rows share the objects in each column.}
\label{tab:appendix-nonrecognition}
\begin{tabularx}{\columnwidth}{@{}Yrr@{}}
\toprule
Source & $\mathcal S_0$ ($300$) & $\mathcal H_0$ ($145$) \\
\midrule
LLaVA-OneVision-1.5-8B & 56.33 & 53.10 \\
Gemini-3-Flash-Preview & 56.00 & 53.79 \\
GPT-4.1-mini & 60.33 & 58.62 \\
Gemini-2.5-Flash-Lite & 57.33 & 54.48 \\
Gemini-2.5-Flash & 57.33 & 53.79 \\
Museum-authored & n/a & 51.72 \\
Original-photograph control & 54.00 & 49.66 \\
\bottomrule
\end{tabularx}
\end{table}

\subsection{Published Recogniser-Selection Results}
Table~\ref{tab:control-current} extracts culture decisions from the five-field outputs published with~\cite{jiang2026appear2meaningcrossculturalbenchmarkstructured}. GPT-4.1-mini is selected by complete coverage followed by culture accuracy: $216/374=57.75\%$. The culture-only control is a separate execution using the reconstruction-recognition prompt, yielding $195/374=52.14\%$ and $122/230=53.04\%$ (Table~\ref{tab:main}). Paired changes and the visualisations above use this separately measured control.

\par\smallskip
\noindent\begin{minipage}{\columnwidth}
\centering\footnotesize
\setlength{\tabcolsep}{2pt}
\captionsetup{hypcap=false}
\captionof{table}{Recogniser selection from the published five-field outputs of~\cite{jiang2026appear2meaningcrossculturalbenchmarkstructured} on the 374-object cohort. The culture field is scored by the DeepSeek-Flash judge; unusable outputs remain unsuccessful in $A/374$. Selection requires $V=374$; bold marks the selected endpoint, which matches $135/230=58.70\%$ on the museum-description subset. Panel (b) reports archived culture matches by region $\times$ type. The separate culture-only control is reported in Table~\ref{tab:main}.}
\label{tab:control-current}
\begin{tabularx}{\columnwidth}{@{}Yrrrrr@{}}
\toprule
\multicolumn{6}{@{}l}{\textbf{(a) All endpoints}} \\
Endpoint & \shortstack{Valid\\$V$} & \shortstack{Culture\\matches $A$} & \shortstack{Valid\\(\%)} & $100A/374$ & $100A/V$ \\
\midrule
Qwen-VL-Max & 369 & 199 & 98.66 & 53.21 & 53.93 \\
Qwen3-VL-Plus & 369 & 216 & 98.66 & 57.75 & 58.54 \\
Qwen3-VL-Flash & 369 & 231 & 98.66 & 61.76 & 62.60 \\
Qwen3-VL-8B & 374 & 164 & 100.00 & 43.85 & 43.85 \\
Qwen3-VL-32B & 374 & 177 & 100.00 & 47.33 & 47.33 \\
Pixtral-12B & 346 & 154 & 92.51 & 41.18 & 44.51 \\
\textbf{GPT-4.1-mini} & \textbf{374} & \textbf{216} & \textbf{100.00} & \textbf{57.75} & \textbf{57.75} \\
GPT-5.4-mini & 374 & 213 & 100.00 & 56.95 & 56.95 \\
Claude Haiku 4.5 & 364 & 193 & 97.33 & 51.60 & 53.02 \\
\bottomrule
\end{tabularx}
\vspace{5pt}
\begin{tabularx}{\columnwidth}{@{}Yrrrrr@{}}
\toprule
\multicolumn{6}{@{}l}{\shortstack[l]{\textbf{(b) Selected recogniser by region $\times$ type:}\\\textbf{correct / total}}} \\
Region & Ceramics & Metalwork & Painting & Sculpture & Overall \\
\midrule
Americas & 7/25 & 13/25 & 13/25 & 8/25 & 41/100 \\
Ancient Mediterranean & 15/25 & 13/25 & 20/25 & 13/25 & 61/100 \\
East Asia & 16/25 & 22/25 & 23/25 & \textit{n/a} & 61/75 \\
Europe & 10/25 & 18/25 & 15/25 & 10/24 & 53/99 \\
\midrule
\textbf{Overall} & 48/100 & 66/100 & 71/100 & 31/74 & 216/374 \\
\bottomrule
\end{tabularx}
\end{minipage}
\par

\par\clearpage
\setlength{\intextsep}{6pt plus 1pt minus 1pt}
\setlength{\textfloatsep}{8pt plus 2pt minus 1pt}
\setlength{\floatsep}{8pt plus 1pt minus 1pt}
\titlespacing*{\paragraph}{0pt}{6pt plus 1pt minus 1pt}{0.6em}

\section{Case Study by Error Type}
\label{sec:case_study}
\label{appendix:error-analysis}
All seven cases are retained from the supplied manuscripts. Each display includes the original image and available candidate passages or reported excerpts, followed by analysis under the supplementary Visual and Culture criteria. Candidate outputs are quoted research material and may contain inaccurate or unsupported claims. The original authors' interpretations are distinguished from directly visible evidence; they are not new independent annotations under the revised rubric. No ablation output or numerical result has been inferred.

\subsection{Case 1: Lion and Text on a Votive Painting}
\label{sec:case1}
\begin{figure}[ht]
\centering
\includegraphics[width=\columnwidth,height=7.5cm,keepaspectratio]{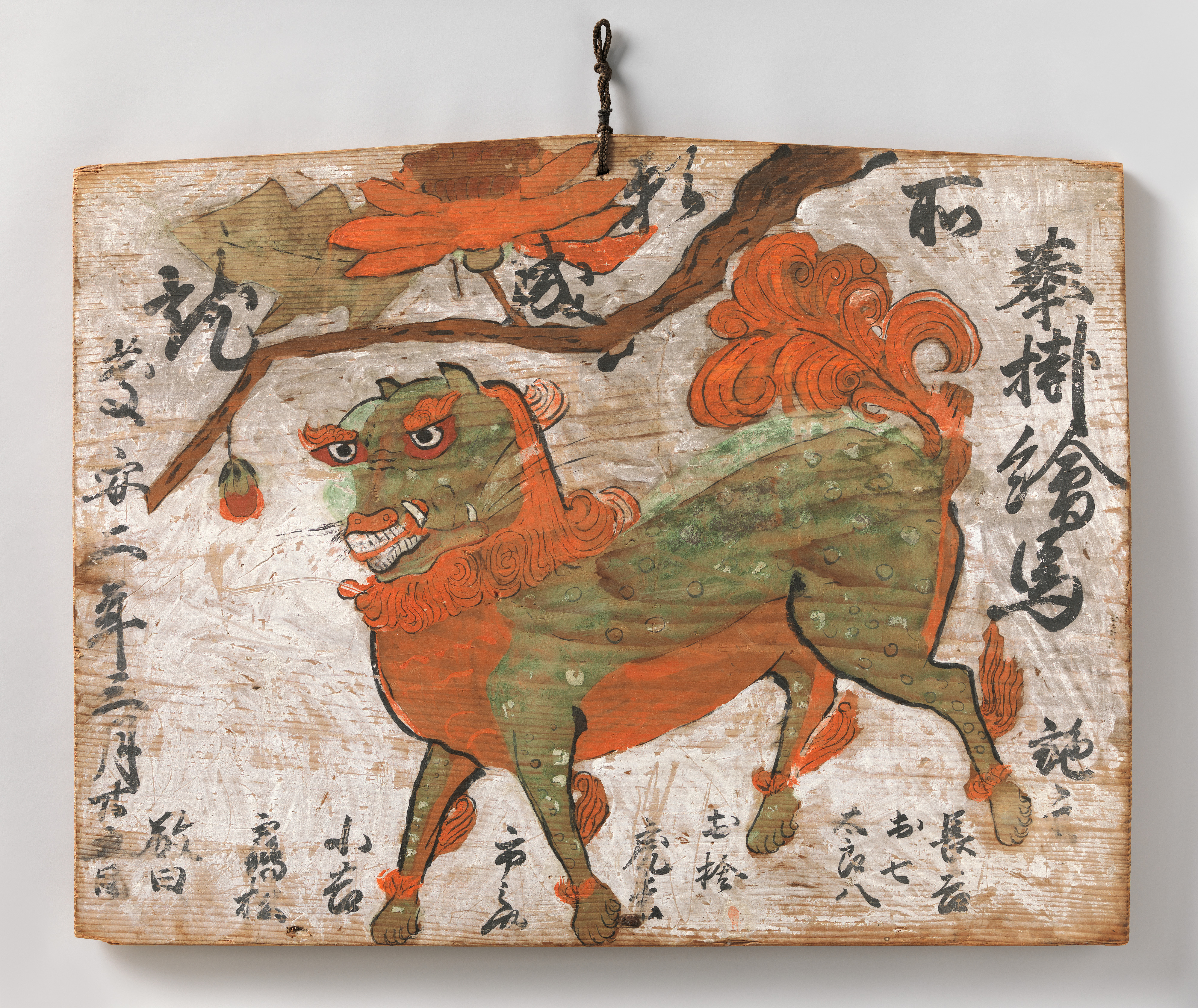}
\caption{Retained artefact image for Case 1.}
\label{fig:case1}
\end{figure}
\paragraph{Original candidate description.}
{\small
This votive painting (ema) from 1649 depicts a stylized Chinese lion beneath a flowering peony tree, rendered in bold orange-red, green, and earth tones on a wooden plaque. The lion—a protective creature in East Asian tradition—is shown in a dynamic, almost playful stance with prominent eyes and an ornamental mane, while the peony blossoms above suggest prosperity and good fortune. Inscribed with Japanese characters around the composition's margins, the work exemplifies the syncretic religious practice of votive offerings at Japanese shrines, where such paintings were hung to petition for blessings or gratitude. The combination of the Chinese lion motif with Japanese calligraphy and the ema format itself reflects the cultural exchange and adaptation of continental Asian imagery within Japanese folk devotional practice during the early Edo period. The painting's modest scale, direct execution, and protective iconography suggest it may have been commissioned by a devotee seeking safeguarding or prosperity for household or family.
}

The first case is a Japanese votive painting dated 1649, with a lion beneath a peony tree and writing around the composition. The candidate discusses protection, good fortune, and ``syncretic religious practice.'' The original readers emphasise the lion together with the inscriptions, particularly the writing on the right.

\paragraph{Visual information.}
The candidate identifies several visible elements and mentions marginal inscriptions. The case concerns how those elements are prioritised: the sustained narrative centres on the lion and peony, while the readers treat text and image as joint anchors. It is therefore a focus discrepancy even where the description contains correct observations. Under the revised rubric, factual accuracy, human-aligned emphasis, and distinguishing inscriptions inform the binary Visual judgement.

\paragraph{Cultural information.}
The candidate's professional vocabulary supplies a general devotional interpretation, but the case notes expect the inscriptions to play a more substantive role in explaining this object. Their observation is not an independent translation of the writing. Without a documented reading, the precise omitted message remains unresolved. The supported qualitative point is that conventional symbolic language occupies the place where an artefact-specific account of the relation between inscription and image is needed.

The case illustrates partial visual description with displaced emphasis and an incomplete cultural explanation. It does not establish that every reference to protection or devotional practice is incorrect, nor that all observers share one visual focus.

\subsection{Case 2: Landscape Album with Trees and Rocks}
\label{sec:case2}
\begin{figure}[ht]
\centering
\includegraphics[width=\columnwidth,height=7.5cm,keepaspectratio]{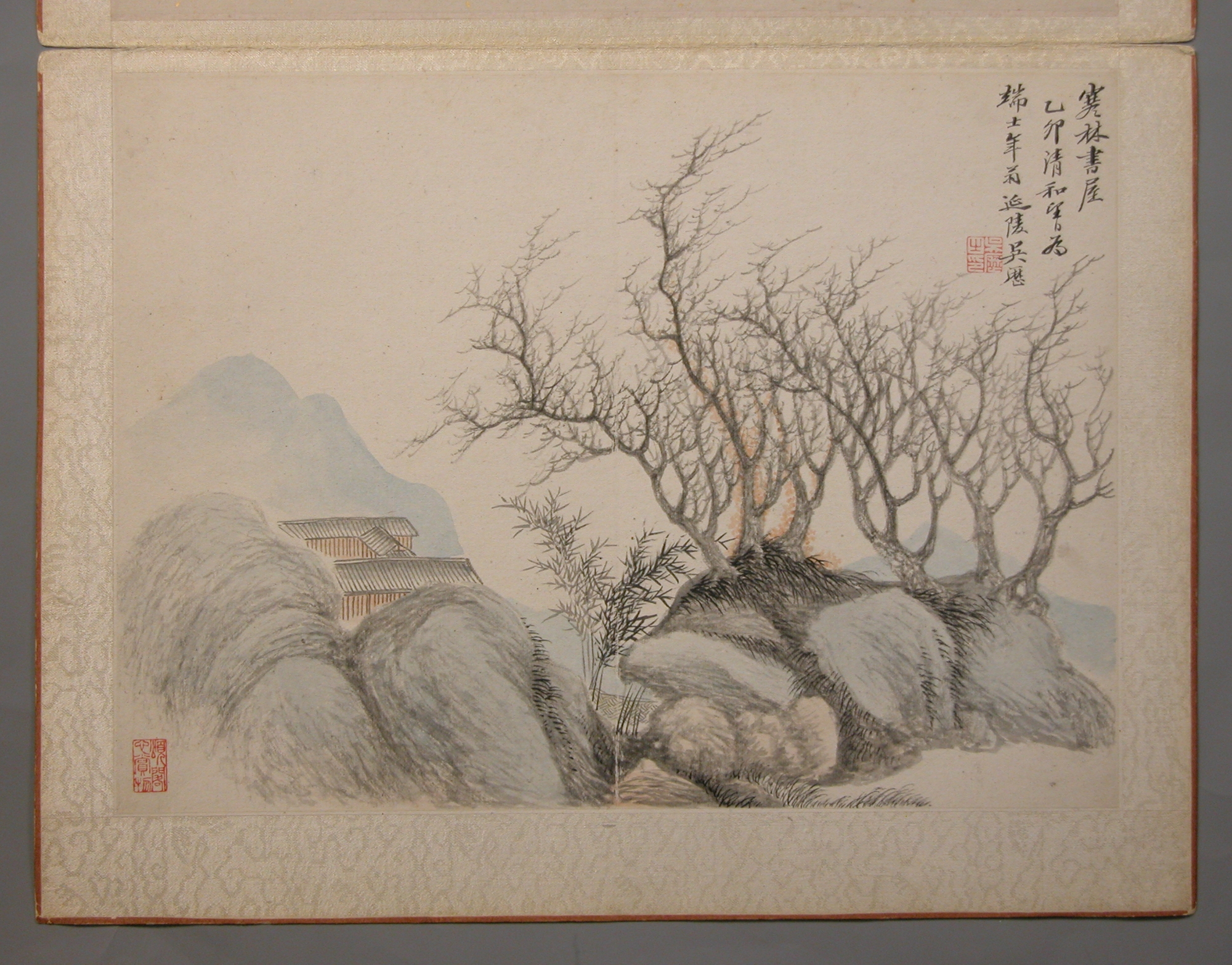}
\caption{Retained artefact image for Case 2.}
\label{fig:case2}
\end{figure}
\paragraph{Original candidate description.}
{\small
Painted album by Wang Jian (1609–1677), a leading literati artist of the early Qing dynasty, created in 1674–76 as a gift for Wang Kui. The album demonstrates the literati painting tradition through a series of landscape compositions rendered in ink and color on paper. Pages alternate between finished landscapes—ranging from intimate scenes of bare trees and scholar's pavilions to expansive mountain vistas with layered peaks, forests, and waterways—and calligraphic inscriptions in running script, accompanied by the artist's seals. The stylistic variety across leaves may suggest Wang Jian's engagement with historical precedents and regional landscape traditions. The gift inscription on the opening leaf indicates this work functioned as a personal offering, a common practice among literati painters for whom artmaking served both aesthetic and social purposes. The album format itself—portable, intimate, and suited to sequential viewing—was particularly valued in literati culture for displaying virtuosity across multiple pictorial themes and demonstrating scholarly refinement through the integration of image and text.
}

The second case is a Chinese landscape album attributed in the supplied material to Wang Jian, dated 1674--76. The candidate invokes ``literati painting tradition'' and ``historical precedents,'' while the original readers foreground the trees and rocks close to the viewer.

\paragraph{Visual information.}
The candidate frames the image through a larger landscape and compositional structure. The readers' account instead treats the foreground elements as the primary visual anchors. This contrast motivates checking human-aligned emphasis within Visual: the issue is not simply whether mountains, trees, or rocks are mentioned, but which elements organise the account. The revised protocol records that priority before catalogue text can influence it.

\paragraph{Cultural information.}
A reference to a literati tradition may be relevant background, yet its relevance to the specific foreground arrangement still requires explanation. The original notes describe a broad art-historical template in place of an account built from concrete visible features. This supports a concern about cultural specificity. It does not by itself show that the historical attribution is false or prove an ontology error merely because the model uses a broad landscape category.

Together with Case~\ref{sec:case1}, this case shows how fluent, correct-sounding vocabulary can coexist with a difference in descriptive emphasis. The evidence concerns the readers' recorded judgements and the candidate text, rather than measured gaze or internal model attention.

\subsection{Case 3: Title-Linked Thematic Emphasis in Calligraphy}
\label{sec:case3}
\begin{figure}[ht]
\centering
\includegraphics[width=\columnwidth,height=7.5cm,keepaspectratio]{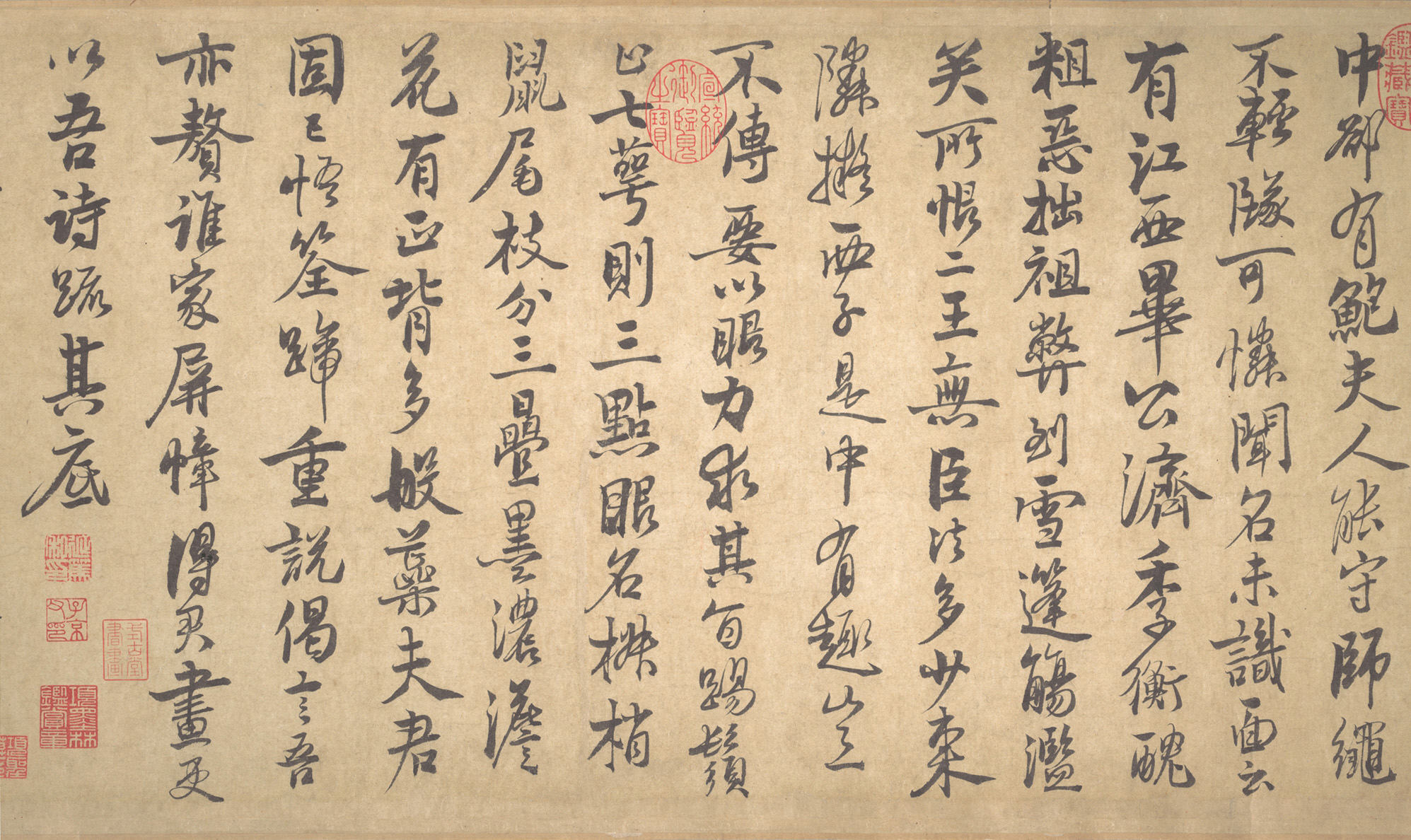}
\caption{Retained artefact image for Case 3.}
\label{fig:case3}
\end{figure}
\paragraph{Original candidate description.}
{\small
A handscroll of poems composed by Zhao Mengjian in 1260, celebrating the subjects of plum blossoms and bamboo through calligraphy. The work demonstrates the literati tradition of linking painting subjects to poetic reflection, with Zhao's bold, confident brushwork rendered in ink on paper. Multiple red seals—both the artist's signatures and later collectors' marks—appear throughout the scroll, indicating its value and circulation among educated patrons. The physical format of the handscroll, with its decorative brocade mounting and protective covering, reflects the refined aesthetic practices of Song-era scholar-artists, for whom the integration of poetry, painting, and calligraphy represented a unified cultural ideal. Plum blossoms and bamboo were emblematic subjects in literati art, valued not merely for their visual appeal but as vehicles for expressing moral character and philosophical virtue.
}

The catalogue title is ``Poems on painting plum blossoms and bamboo.'' The candidate explicitly identifies a handscroll of poems and calligraphy, then elaborates the symbolism of the plants named in the title. The original readers foreground the visible writing and seals. The surviving passage supports a difference in descriptive emphasis; it does not explicitly assert that plants are painted on the object.

\paragraph{Visual information.}
The technical concern is which information organises the account. Although the candidate mentions calligraphy, brushwork, and seals, its extended thematic discussion may give less prominence to the visible textual composition than the original readers expect. This is their recorded focus judgement, not a new primary-stage rating or evidence that the model failed to recognise the object as calligraphy.

\paragraph{Cultural information.}
The candidate links poetry, painting subjects, and plant symbolism. Assessing the relevance of those interpretations requires sources for this particular handscroll. A title-linked theme can be pertinent context while leaving the object's distinctive visual features underdescribed. The quoted output alone does not establish that a poem about painting was mistaken for a painting of plants.

The title--output correspondence motivates examining input dependence, but does not establish a causal title effect or an internal attention pattern. No paired input-removal outputs are available for this case.

\subsection{Case 4: A Wall Painting with a Disputed Cultural Frame}
\label{sec:case4}
\begin{figure}[ht]
\centering
\includegraphics[width=\columnwidth,height=7.5cm,keepaspectratio]{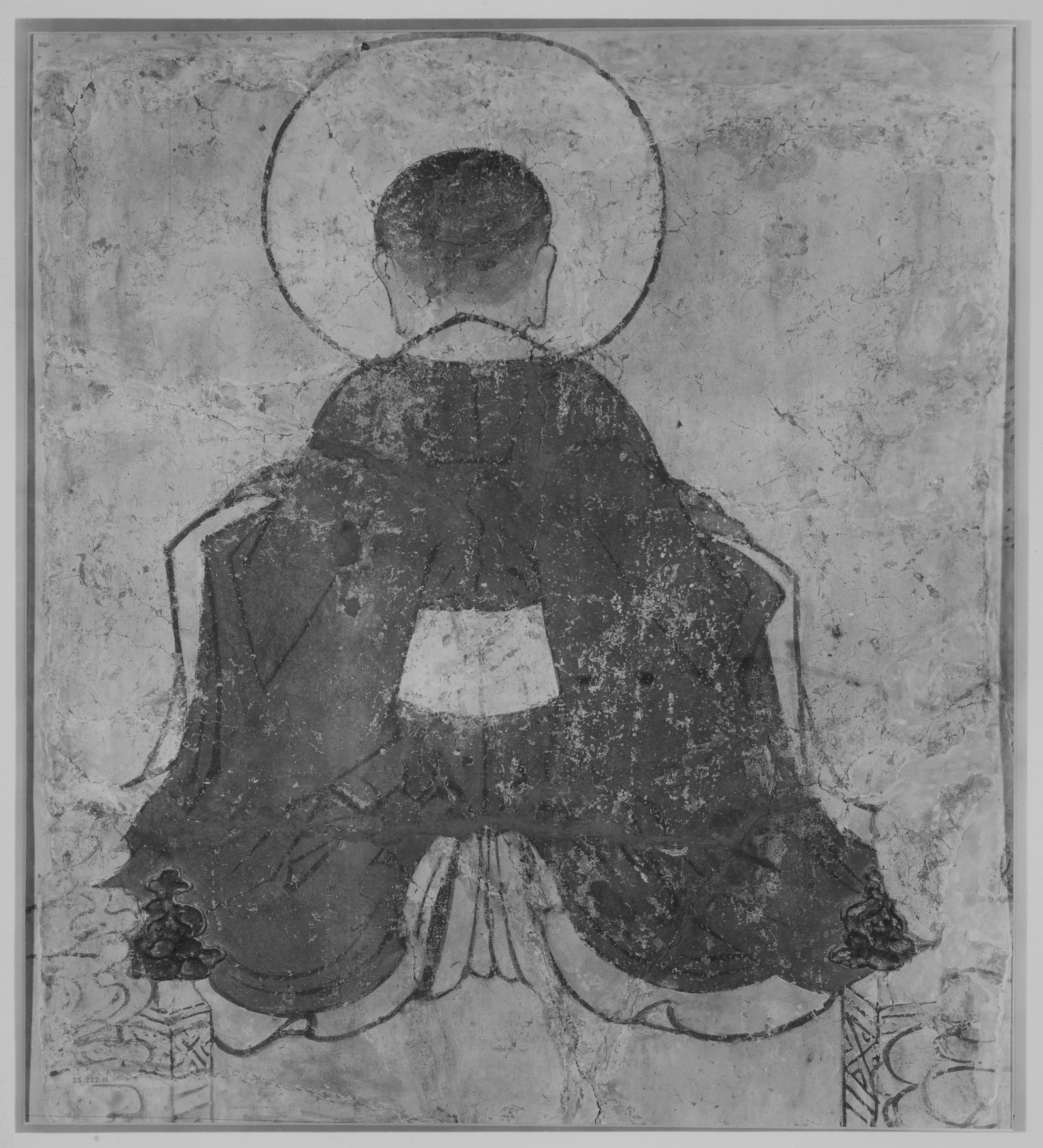}
\caption{Retained artefact image for Case 4.}
\label{fig:case4}
\end{figure}
\paragraph{Original candidate description.}
{\small
A Chinese wall painting depicting a haloed figure in formal robes, rendered in dark pigment against a light plaster ground. The figure is shown in frontal pose wearing layered garments with white rectangular panels at chest and waist, characteristic of official or ceremonial dress. A prominent circular halo frames the head, a motif typically associated with religious veneration or divine status in Chinese artistic traditions. Small inscriptions or decorative elements appear in the lower corners. The work's austere compositional style, emphatic use of the halo, and formal iconography suggest this may represent a revered religious, ancestral, or imperial figure. The painting's weathered surface and accumulated patina indicate considerable age, though the precise period and original architectural context remain uncertain. The work exemplifies the tradition of devotional or commemorative wall painting in Chinese religious and domestic spaces.
}

The fourth case is a Chinese wall painting. The original readers describe a figure facing away from the viewer and a railing in front; they interpret the figure as a back-facing Buddha. The candidate gives a frontal account and uses general religious and material vocabulary, including ``formal robes,'' ``halo,'' and ``dark pigment.''

\paragraph{Visual information.}
The orientation of the figure and the distinction between railing and inscription are the principal image-checkable concerns. The original notes treat the back-facing pose as a defining feature, while the candidate foregrounds clothing and conventional religious symbols. The revised analysis separates these visual errors from any claim that the pose is historically rare, which requires additional documentary evidence.

\paragraph{Cultural information.}
The readers expect an explanation of the back-facing pose and its relation to the figure's identity. Their objections to the English vocabulary are preserved as situated interpretive concerns. However, ``halo'' is not intrinsically Western or inappropriate for a Buddhist image, and ``formal robes'' or ``dark pigment'' alone do not establish colonial framing. A terminology judgement must identify a misleading use in context. Likewise, replacing pigment terminology with ink or brushwork requires evidence about this object's medium.

The supported concern is that a general religious narrative accompanies an inaccurate or incomplete account of the particular figure. The case motivates object-specific cultural explanation and contextual review of vocabulary, rather than an automatic ban on individual words.

\subsection{Case 5: Goldfish in a Glass Bottle}
\label{sec:case5}
\begin{figure}[ht]
\centering
\includegraphics[width=\columnwidth,height=7.5cm,keepaspectratio]{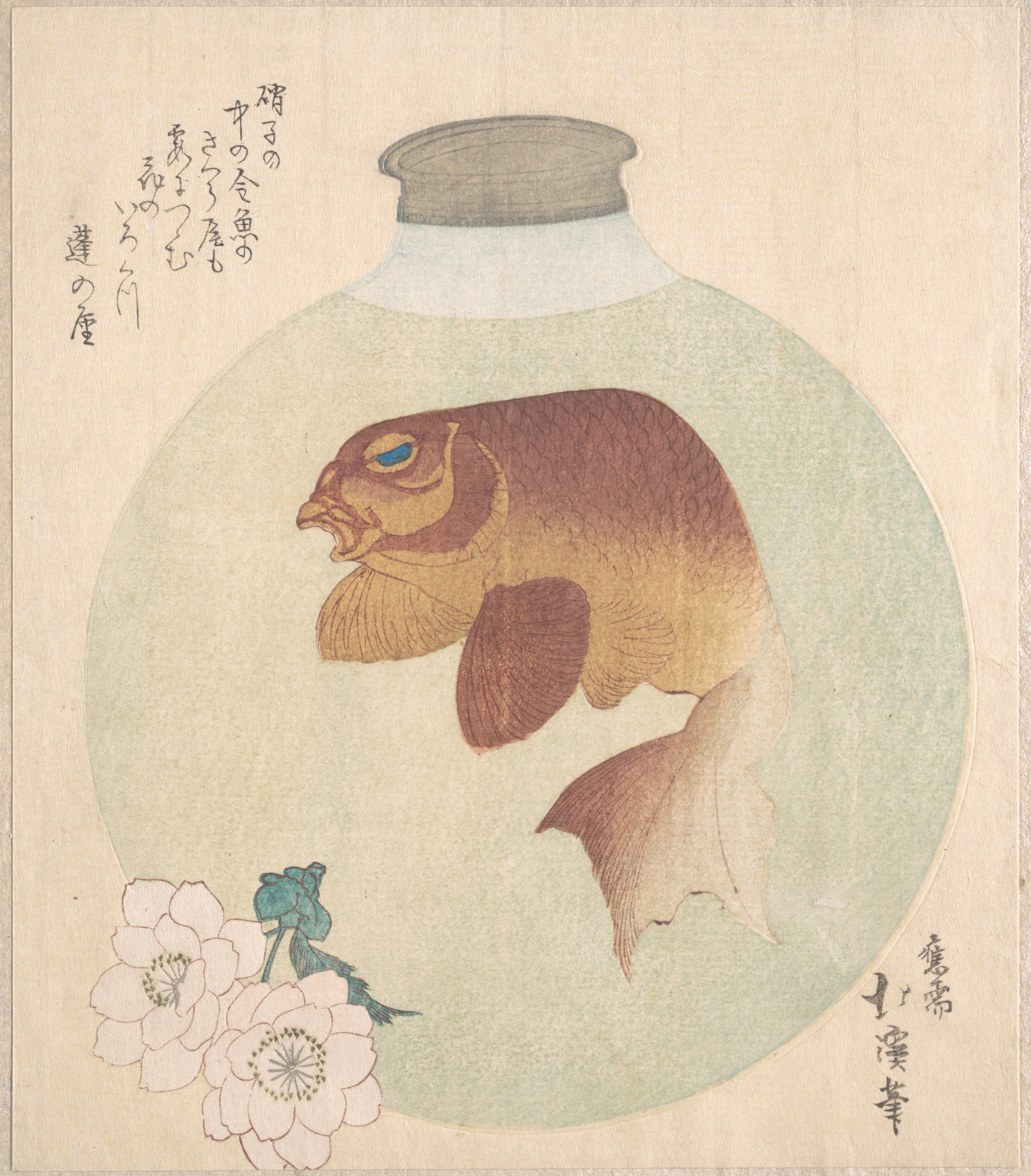}
\caption{Retained artefact image for Case 5.}
\label{fig:case5}
\end{figure}
\paragraph{Original candidate description.}
{\small
This woodblock print by Totoya Hokkei depicts a goldfish suspended in a glass bottle filled with water—a composition that transforms a domestic object into a study of transparency, containment, and natural observation. The fish, rendered in warm ochres and deep reds with careful attention to scale and fin detail, appears almost paradoxically lifelike within the geometric clarity of its vessel. The bottle's glass is suggested through subtle gradations of tone rather than outline, demonstrating the technical precision characteristic of Japanese ukiyo-e printing in the 19th century. A spray of delicate peonies in the lower left provides a counterpoint to the contained subject, while calligraphic inscriptions frame the composition. The work reflects the period's fascination with depicting transparent and reflective surfaces, and may suggest the influence of Western representational techniques on Japanese printmaking, while remaining grounded in the tradition of nature study (surimono) that valued intimate observation of small, domestic subjects. The goldfish—historically associated with leisure, refinement, and the natural world brought indoors—becomes here a meditation on the act of seeing itself.
}

The fifth case is a Japanese woodblock print showing a goldfish in a glass bottle. The candidate expands the scene into a discussion of observation, transparency, and Western influence, using complex wording such as ``paradoxically.'' The original readers describe the composition as visually direct and find the extended interpretation difficult to assess.

\paragraph{Visual information.}
The visible fish, container, and simple arrangement provide the readers' main anchors. The candidate devotes greater emphasis to abstract concepts. This is a Visual concern when the conceptual account displaces the depicted scene; sentence length or sophistication alone is not a visual error.

\paragraph{Cultural information.}
The central issue is the support for the proposed historical influence and deeper interpretation. The supplied case notes identify no clear object-specific evidence for a Western-influence claim. Such a claim may be assessed through documentation, but it is not established by glass transparency or by the use of a familiar interpretive vocabulary. For Culture, abstract ideas must explain this artefact and historical interpretation requires supporting evidence or appropriate qualification.

The case illustrates overinterpretation rather than a universal incompatibility between complex prose and Japanese art. The word ``paradoxically'' is not a cultural-bias indicator. Even an output without a vocabulary-list hit can require human review because unsupported framing is a sentence-level issue.

\subsection{Case 6: Material Attribution in a Bird-and-Flower Print}
\label{sec:case6}
\begin{figure}[ht]
\centering
\includegraphics[width=\columnwidth,height=7.5cm,keepaspectratio]{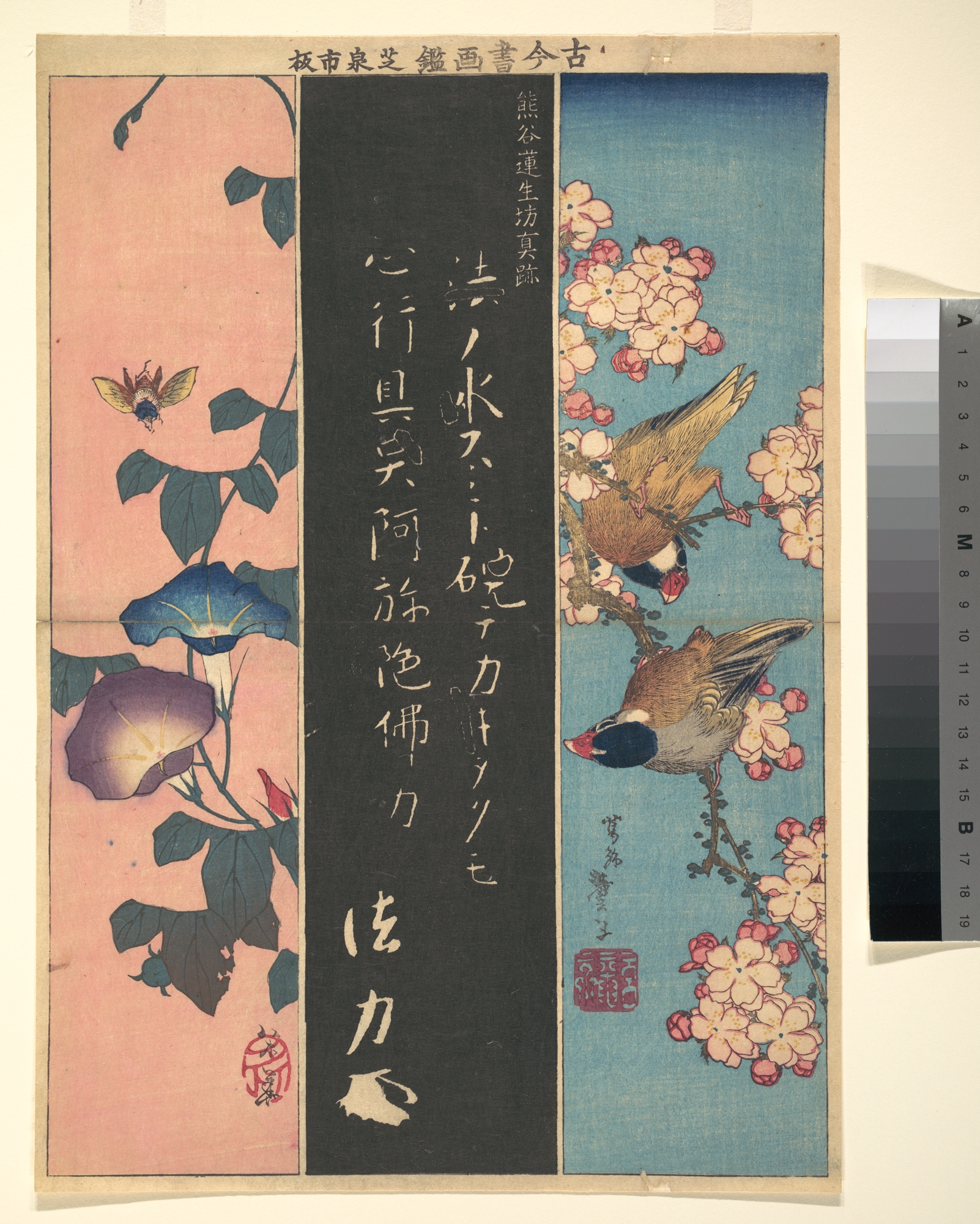}
\caption{Retained artefact image for Case 6.}
\label{fig:case6}
\end{figure}
\paragraph{Recorded candidate excerpts.}
The complete, unedited candidate is not preserved in the supplied record. The case notes report ``Japanese text in gold,'' ``small white and coral blossoms,'' ``decorative refinement,'' and ``observational sensitivity.'' The following discussion relies on those reported excerpts and the readers' interpretations, not a reconstructed full response.

The sixth case is a Japanese bird-and-flower print attributed in the supplied notes to Katsushika Hokusai. The original readers instead interpret the yellow tone as paper ageing and associate the blossoms with cultural symbolism.

\paragraph{Visual information.}
The critical distinction is between visible yellow colour and an assertion about its material cause. Neither gold pigment nor ageing is established solely by a colour observation. The reported material attribution and the readers' alternative both require catalogue or conservation support. The notes motivate checking that proposition; they settle neither explanation and provide no new primary-stage rating.

\paragraph{Cultural information.}
The original notes argue that generic colour-and-shape descriptions flatten the symbolic significance of flowers and omit material history. Their identification of cherry blossoms and expected symbolic reading are retained as interpretations requiring an artefact-specific source. The same applies to their cross-case analogy with peonies: a flower name alone is not a complete or universal cultural explanation.

This case exposes the difference between naming visual attributes and explaining their historically supported significance. It supports scrutiny of unsupported material claims and underdeveloped cultural context, without treating the readers' preferred botanical or conservation attribution as independently verified fact.

\subsection{Case 7: Natural Stone Patterns and Human Selection}
\label{sec:case7}
\begin{figure}[ht]
\centering
\includegraphics[width=\columnwidth,height=7.5cm,keepaspectratio]{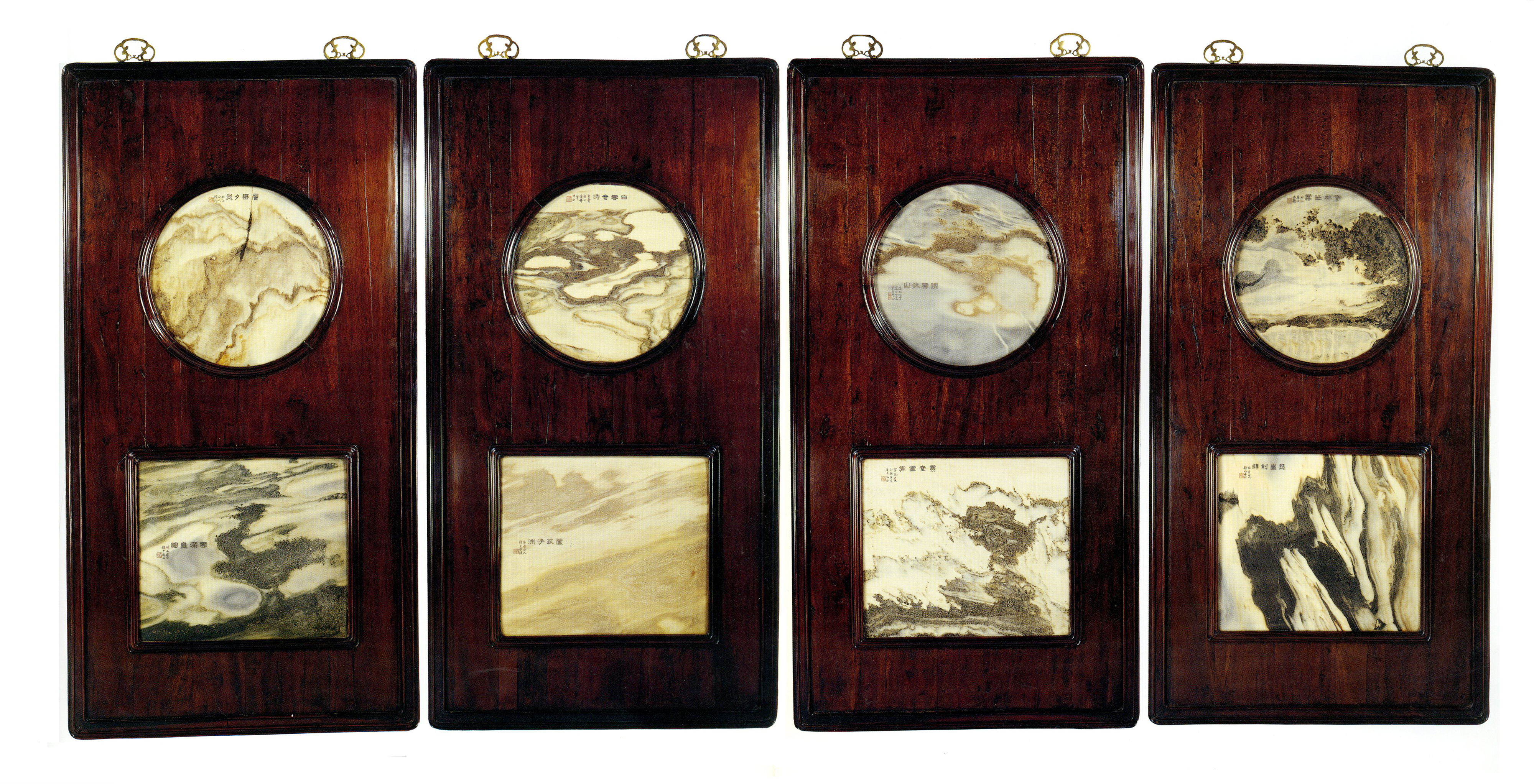}
\caption{Retained artefact image for Case 7.}
\label{fig:case7}
\end{figure}
\paragraph{Original candidate description.}
{\small
Set of four 19th-century Chinese wall panels featuring stone paintings mounted on polished hardwood. Each panel is organized with a circular composition at the top and a rectangular landscape below, both executed in ink on stone with subtle tonal gradations. The paintings depict characteristic Chinese landscape motifs—misty mountains, flowing rivers, and weathered rock formations rendered in the literati painting tradition. Chinese characters inscribed on each work suggest titles or artist seals. The deep reddish-brown wood frames, fitted with brass hanging loops, provide an austere backdrop that emphasizes the luminous quality of the stone surfaces. This presentation format reflects 19th-century scholar-collector practices, where paintings on unconventional materials such as stone were prized for their material durability and aesthetic distinctiveness. The ensemble demonstrates how landscape imagery—a cornerstone of Chinese artistic philosophy—was adapted to stone as both a technical and conceptual challenge, suggesting the paintings may have served a decorative or contemplative function within an educated household or scholarly setting.
}

The seventh case concerns Chinese wall panels described in the supplied material as stone paintings. Natural stone slabs were selected for patterns resembling landscapes, then cut and mounted. The original interpretation centres on the relation between natural formation and human selection. The candidate instead uses familiar painting vocabulary, including ``literati painting tradition,'' ``landscape motifs,'' and ``scholar-collector practices.''

\paragraph{Visual information.}
The candidate treats natural textures as conventional painted scenery. Where the material record establishes patterned stone, an account of brushstrokes mischaracterises the object. The readers foreground how the panels organise natural patterns; the candidate foregrounds a standard landscape composition. Identifying the medium may require the catalogue record in addition to the photograph, so this case does not isolate pure perception from domain knowledge.

\paragraph{Cultural information.}
The meaningful distinction is between a landscape created through painting and a landscape recognised in a natural material and framed through human selection. General art-historical phrases can be compatible with appreciation of the object while still leaving that distinction unexplained. The original notes describe the work as nature-made and human-curated; this is an interpretive account of making and selection, rather than proof of a universal theory of authorship.

The case illustrates how a familiar cultural category can simplify the object's particular logic. It is consistent with metadata-associated framing but cannot on its own distinguish a textual shortcut from inadequate material recognition or specialist knowledge. Culture requires an object-linked explanation, while Visual requires accurate, prominent, distinguishing features; neither infers an internal model representation.

\section{Supporting Qualitative Evidence}
\label{appendix:legacy-experiment-notes}
\subsection{Getty Reconstruction Pilot}
\label{sec:getty-pilot}
The historical description-only pilot retains twelve Getty descriptions, their source objects and an LLM-assisted editorial fluency screen. One image request per description yielded nine images and three tool-blocked executions: \emph{Night}, \emph{Statuette of an Amazon}, and \emph{Saturn Devouring One of His Children}. Requests were not retried. The pilot archive preserves the source text, fixed rendering prompts, images, execution statuses and comparison notes. This record is separate from the 374-object experiment. Earlier three-panel Getty illustrations use metadata plus descriptions in a different execution; their provenance remains attached to that archive.

\begin{table}[htbp]
\centering\footnotesize
\setlength{\tabcolsep}{3pt}
\caption{Diagnostic distinctions in the historical Getty comparison notes. Each observation is tied to the saved description and source image.}
\label{tab:getty-diagnostic-notes}
\begin{tabularx}{\columnwidth}{@{}p{0.21\columnwidth}YY@{}}
\toprule
Example & Recorded discrepancy & Diagnostic question \\
\midrule
Wall-light & Text specifies a classical mask and symmetric arms; source objects have animal heads and asymmetric branches. & Does the description explicitly specify a different structure? \\
Jar & The generated portrait faces the opposite way; text leaves direction unspecified. & Which spatial relation is left open in the description? \\
Vases & Reconstruction includes lids absent from the source photograph. & Which visible details are completed during generation? \\
\bottomrule
\end{tabularx}
\end{table}

Table~\ref{tab:getty-diagnostic-notes} distinguishes explicit misdescription, underspecified spatial information and generator completion. Reading the unedited text beside both images identifies where a discrepancy enters the chain. These are recorded LLM-assisted qualitative observations with their own execution provenance.

\subsection{Supplementary Review Criteria}
The retained cases in Appendix~\ref{sec:case_study} organise observations under two criteria. \emph{Visual} concerns the accuracy and prominence of distinguishing visible features. \emph{Culture} concerns how an object-specific interpretation connects those features to documented cultural, material or historical relations. The case notes preserve the original readers' interpretations and distinguish full candidate passages from reported excerpts.

Across the cases, these criteria locate three recurring issues. First, mentioning a feature can leave it peripheral to the account: inscriptions or foreground trees and rocks may appear in the text while a conventional thematic narrative organises it. Second, a visible attribute can become an evidentially stronger claim, as when yellow colour is described as gold or a pose is assigned a specific orientation. Third, a familiar art-historical account can replace the object's material logic, as in the distinction between painted scenery and naturally patterned stone selected and mounted by people. Catalogue and material evidence determine which proposed interpretation is supported in each case. The complete passages, images and case-specific discussion remain in Appendix~\ref{sec:case_study}.

\end{document}